\documentclass[preprint,12pt]{elsarticle}

\usepackage[T1]{fontenc}
\usepackage[utf8]{inputenc}
\usepackage{lmodern}
\usepackage{microtype}

\usepackage{amsmath,amssymb,mathtools,bm}
\usepackage{physics}
\usepackage{braket}
\usepackage{cancel}
\usepackage{simpler-wick}

\usepackage{graphicx}
\usepackage{subcaption}
\usepackage{svg}
\usepackage{siunitx}
\usepackage{multirow}
\usepackage{booktabs}
\usepackage{threeparttable}
\usepackage{tablefootnote}
\usepackage{float}

\usepackage{enumerate}
\usepackage{tikz}
\usetikzlibrary{decorations.pathreplacing}

\usepackage[hidelinks]{hyperref}

\journal{Engineering Applications of Artificial Intelligence}

\begin{document}

\begin{frontmatter}


\title{Recovering Sharp Conductivity Features in the Finite-Data Calder\'on Problem with Physics-Informed Neural Networks}

\author[iccub,dept]{Ali AlHadi Kalout}
\author[iccub,dept]{Pablo Tejerina-P\'erez}
\author[iccub]{Konstantin Karchev}
\author[iccub,dept]{Pedro Taranc\'on-\'Alvarez}
\author[iccub,dept]{Leonid Sarieddine}
\author[iccub,icrea]{Raul Jimenez}
\author[umn]{Max Engelstein}
\author[orsay]{Guy David}

\affiliation[iccub]{organization={Institut de Ci\`encies del Cosmos (ICCUB), Universitat de Barcelona},
  addressline={Mart\'i i Franqu\`es 1},
  postcode={ES-08028},
  city={Barcelona},
  country={Spain}}

\affiliation[dept]{organization={Departament de F\'isica Qu\`antica i Astrof\'isica, Universitat de Barcelona},
  addressline={Mart\'i i Franqu\`es 1},
  postcode={ES-08028},
  city={Barcelona},
  country={Spain}}

\affiliation[icrea]{organization={ICREA},
  addressline={Pg. Lluis Companys 23},
  postcode={08010},
  city={Barcelona},
  country={Spain}}

\affiliation[umn]{organization={School of Mathematics, University of Minnesota},
  city={Minneapolis},
  postcode={MN 55455},
  country={USA}}

\affiliation[orsay]{organization={Universit\'e Paris-Saclay, Laboratoire de Math\'ematiques d'Orsay},
  postcode={91405},
  country={France}}

\begin{abstract}  
Physics-informed neural networks (PINNs) have recently emerged as a promising framework for addressing the Calder\'on inverse problem from limited boundary data. In this work, we revisit neural Calder\'on inversion by introducing multiscale boundary excitations based on randomized wavelet functions and investigating the role of Fourier-feature encoding (FFE) for representing sharp conductivity variations. We propose a physics-informed reconstruction framework that represents the unknown conductivity and the associated family of electric potentials with separate neural networks conditioned on the applied boundary excitations. The governing elliptic PDE is enforced through physics-informed residuals, while finite Dirichlet-to-Neumann (DtN) data are incorporated through boundary losses. Using synthetic data from a finite-difference forward solver, we evaluate the method on conductivity fields with inclusions, sharp interfaces, smooth profiles, and heterogeneous media. Results show that the framework recovers dominant conductivity structures from finite boundary measurements with relative errors between $3\%-12\%$ approximately. We show that FFE improves the reconstruction of localized sharp features, particularly for inclusions and interfaces, but are not universally optimal, with raw-coordinate networks performing competitively for smoother fields. These results highlight coordinate representations and boundary excitation design as key factors in neural Calder\'on inversion.
\end{abstract}


\end{frontmatter}

\section{Introduction}

Inverse problems play a central role across physics, engineering, and applied mathematics, as they aim to infer hidden properties of a system from indirect or incomplete observations. In contrast to forward problems---where model parameters are known and observable quantities are computed---inverse problems require reconstructing unknown coefficients, sources, or geometrical features from measured data. Such problems are typically ill-posed in the sense of Hadamard, lacking uniqueness, stability, or continuous dependence on the data. Consequently, their solution requires careful mathematical formulation and robust numerical methodologies.

A paradigmatic example of an inverse problem is the \emph{Calder\'on problem}, originally introduced by Calder\'on in 1980 \cite{calderon_original_reprint} in the context of electrical impedance tomography (EIT). The problem consists of determining the spatially varying electrical conductivity $\gamma(x)$ of a medium from boundary measurements of voltage and current. Formally, given a bounded domain $\Omega \subset \mathbb{R}^n$, the electric potential $u(x)$ satisfies the elliptic partial differential equation
\begin{equation}
\notag
\nabla \cdot (\gamma(x)\nabla u(x)) = 0,
\end{equation}
subject to Dirichlet boundary conditions (BCs) $u|_{\partial\Omega} = f$, with $f$ being the electric potential at the boundary of the domain. The associated Dirichlet to Neumann (DtN) map,
\begin{equation}
\notag
\Lambda_\gamma(f) = \left.\gamma(x)\frac{\partial u}{\partial \hat{n}}\right|_{\partial\Omega},
\end{equation}
encodes the boundary response of the system, i.e. the induced electric current at the boundary given an electric potential $f$. The inverse Calder\'on problem then consists of reconstructing $\gamma(x)$ from knowledge of $\Lambda_\gamma$. This problem has deep connections to medical imaging, geophysical exploration, and non-destructive testing, where one seeks to identify internal structures---such as tumors or inclusions---from boundary measurements alone.

From a theoretical standpoint, the Calder\'on problem has been extensively studied over the past decades. Foundational results have established uniqueness of the reconstruction under various regularity assumptions \cite{SylvesterUhlmann1987}, as well as stability estimates that are typically logarithmic in nature \cite{Alessandrini1988StableDO}, reflecting the severe ill-posedness of the problem. Despite these advances, practical reconstruction remains highly challenging, particularly in settings with limited or noisy data, finite boundary measurements, or complex geometries. Classical numerical approaches include optimization-based methods, regularization techniques, and iterative solvers, but these often suffer from high computational cost and sensitivity to initialization and noise \cite{Mueller_Dbar}.

In recent years, machine-learning methods have provided new numerical tools for inverse problems governed by PDEs. Physics-Informed Neural Networks (PINNs) are particularly relevant in this context because they incorporate the governing equations and BCs directly into the training objective, allowing unknown fields and PDE coefficients to be inferred without requiring a fully labeled training dataset, see \cite{RAISSI2019686, yang2021bayesian, Bea:2024xgv, tarancon2025efficient} among others. In EIT and the Calder\'on problem, it is important to distinguish between two learning settings. In a \emph{semi-inverse} setting, the electric potential inside the domain is assumed known, whereas in the \emph{full-inverse} setting only boundary measurements are available, making the problem significantly more ill-posed.

A first relevant PINN-based approach to PDE-based tomography was proposed by Bar and Sochen \cite{BarSochen2021}. Their method represents the unknown fields by neural networks (NNs) and trains them in an unsupervised manner. In the EIT setting, they considered forward, semi-inverse, and full tomography formulations. In their full tomography formulation, several potential networks, one for each imposed current pattern, are trained simultaneously with a conductivity network. This makes their work conceptually close to ours: both approaches attempt to recover a conductivity field by jointly enforcing the PDE and boundary measurements. However, their implementation uses separate networks for the different measurements, works mainly with smooth conductivity phantoms obtained by Gaussian smoothing, and uses trigonometric current patterns on circular or free-form domains.

Pokkunuru et al.~\cite{Pokkunuru2023} addressed training instability by introducing a Bayesian energy-based prior over conductivities. While this improves robustness, their main focus remains the semi-inverse setting, where interior potential information is available. In contrast, our method operates purely from boundary data without imposing a learned prior.

More recently, Yang et al. \cite{Yang2024CPFI} proposed a hybrid CNN--PINN framework that decouples the inverse problem: a supervised CNN (Convolutional Neural Network) first predicts interior potentials from boundary measurements, and a PINN then reconstructs the conductivity. Unlike this approach, we do not rely on supervised learning or precomputed datasets of conductivity--measurement pairs, but instead solve each inverse problem instance directly.

Operator-learning methods provide a complementary route to the inverse Calder\'on problem. DeepONet-based theoretical results have shown that neural operators can approximate mappings related to the Calder\'on problem~\cite{Castro2024,DeepONets}, while Neural Inverse Operators combining DeepONets and Fourier Neural Operators have achieved high accuracy in supervised reconstruction settings~\cite{molinaro2023neuralinverseoperators,li2021fourierneuraloperator}. These approaches learn an inverse map from a distribution of training examples. In contrast, our method does not train a global inverse operator over a dataset of conductivities, but instead solves each inverse problem instance directly through the PDE and the available finite boundary measurements.

In this work, we study a PINN-based approach to the full Calder\'on inverse problem on a two-dimensional square domain. The method uses two coupled NNs: a conductivity network representing \(\gamma(x,y)\), and a single shared potential network representing the family of solutions \(u_k(x,y)\), where $k = \{1,\dots, K\}$ represents different boundary excitations. The potential network is conditioned on the BC index through a one-hot encoding, allowing one network to represent all potentials associated with the finite set of boundary excitations. The conductivity network is common to all BCs, enforcing the fact that the same underlying conductivity must explain all measured boundary responses.

A central component of this work is the role of coordinate representation in PINN-based inversion. Standard coordinate-based multilayer perceptrons are known to exhibit a \textit{spectral bias} \cite{spectral_bias_Rahaman_2019}, tending to learn low-frequency components more easily than high-frequency ones. This can lead to overly smooth reconstructions, especially for discontinuous or sharply localized conductivities. We therefore compare raw-coordinate networks with Fourier-feature encoded networks in order to assess whether enriched coordinate representations improve the recovery of inclusions, interfaces, oscillatory structures, and heterogeneous profiles with randomly (gaussian) distributed frequency components.

A second feature of the method is the use of randomized wavelet-based BCs. Since only finitely many measurements are available, the choice of boundary excitations plays an important role in the practical identifiability of the conductivity. Rather than using simple trigonometric or electrode-based patterns, we use localized multiscale wavelet superpositions designed to probe the domain from different boundary locations and spatial scales.

The structure of the paper is as follows. In Section \ref{sec 2: theory} we review the theoretical formulation of the Calder\'on problem and its mathematical properties. Section \ref{sec 3: methodology} presents the methodology, including the construction of the synthetic dataset, the choice of BCs, and the PINN architecture used for inversion. In Section \ref{sec: results} we present and discuss the results for different classes of conductivity profiles, analyze the role of Fourier Feature Encoding in the reconstruction performance, and quantify the sensitivity of our method to the more internal regions of the domain, where the instability properties are known to be most relevant. Finally, in Section \ref{sec 6: conclusions} we discuss the implications of our findings and outline directions for future work. This work also includes three appendices.~\ref{appendix:fdm solver} presents the conventional forward numerical solver used in our analysis to generate synthetic data.~\ref{ap:conductivity_profiles} provides the explicit expressions for the ground-truth conductivity profiles. Finally,~\ref{appendix: training and loss specifications} details the neural network architecture together with the training and loss specifications.

\section{Theoretical Framework of the Calder\'on Problem}
\label{sec 2: theory}

Let $\Omega \subseteq \mathbb{R}^n$ be a bounded domain with smooth boundary\footnote{There exist many results where this condition can be relaxed to domains with Lipschitz boundaries \cite{Uhlmann_2009}.}, and let the electric conductivity of the medium spanning the domain be a positive, bounded, real function $\gamma(x)$, where $x\in \Omega$. The induced electric potential $u(x)\in H^1(\Omega)$ (one derivative in $L^2$) in the domain will satisfy, in absence of sources, the continuity equation \cite{Uhlmann_2009}:
\begin{equation}
\label{eq: calderon ODE}
    \nabla\cdot\left(\gamma(x)\,\nabla u(x)\right) = 0\,\,,
\end{equation}
where $\gamma(x)\nabla u(x)$ is the electric flux. Given an electric potential $f\in~H^{1/2}(\partial\Omega)$ (one-half derivative in $L^2$) on the boundary of the domain $\partial\Omega$, then the Dirichlet problem is defined as:
\begin{equation}
\label{eq: calderon ODE + BC}
    \left\{ \begin{array}{l}
     \nabla\cdot\left(\gamma(x)\,\nabla u(x)\right) = 0 \\ \\
     u(x)|_{\partial\Omega} = f\,\,.
    \end{array}\right.
\end{equation}

This is an elliptic PDE with Dirichlet BCs. One can then define a measurement performed on the boundary which maps an electric potential $f\in H^{1/2}(\partial\Omega)$ to an electric current $J  \in H^{-1/2}(\partial \Omega)$. This map from $f$ to $J$ is known as the Dirichlet-to-Neumann (DtN) map (or voltage-to-current map) and we denote it $\Lambda_\gamma$ so that:
\begin{equation}
\label{eq: dirichlet to neumann map}
    \Lambda_\gamma(f) = \left[ \gamma(x)\frac{\partial u}{\partial \hat{n}} \right]_{\partial\Omega}\equiv J \,\,,
\end{equation}
where $\hat{n}$ represents the unit outer normal vector to $\partial\Omega$. The inverse Calder\'on Problem consists of determining $\gamma(x)$ given the DtN map $\Lambda_\gamma$ \cite{Uhlmann_2009}. There are proofs in 2D and higher dimensions of the bijectivity of this relation. Uniqueness results for isotropic scalar conductivities are known in both higher dimensions and in two dimensions under different regularity assumptions~\cite{SylvesterUhlmann1987,Nachman1996,AstalaPaivarinta2006, CARO_ROGERS_2016}; see~\cite{Uhlmann_2009} for a review.

Despite these results on the solvability of the Calder\'on problem, it is well understood that the problem is highly unstable. Foundational work of Alessandrini \cite{Alessandrini1988StableDO} and later Mandache \cite{Mandache2001} shows that even in the smooth setting one cannot expect better than log-stability; that is to say a perturbation of size $\approx 1/k$ in the conductivity can result in a difference in observed measurements on the order of $\approx 10^{-k}$. However, better stability can occur
when stronger assumptions on the algebraic structure or symmetries of the conductivities are made {\it a priori}, see the survey, \cite{Alessandrini_survey_2007}, for more details.

A useful physical interpretation of the problem is its application to tumor detection in the brain using electrical measurements, a procedure known as \textit{electric impedance tomography} (EIT). Here, the brain is modeled as the bounded domain $\Omega$, and the scalp as its boundary $\partial\Omega$. The internal, spatially varying conductivity $\gamma(x)$ changes with the electrical properties of healthy and tumorous tissue. By applying different electric potentials $f_k$ on the scalp and measuring the resulting boundary currents $J_k$, one aims to infer the location and size of the tumor by recovering the full electric conductivity of the brain. Since a single measurement is insufficient, multiple boundary excitations and corresponding current measurements (labeled by $k=1,\dots, K$, where $K$ denotes the total number of measurements) are required to accurately reconstruct the conductivity distribution through the DtN map $\Lambda_\gamma$.

This analogy illustrates a practical limitation: in realistic settings, one cannot perform an arbitrarily large number of measurements. For this reason, throughout this work we restrict our attention to the so-called ``reduced DtN map'', which consists of a finite set of boundary measurements $\{f_k, J_k\}_{k=1}^{K}$. Strictly speaking, such a finite-data setting can not guarantee uniqueness on its own. Consequently, the recovered conductivity should be interpreted as one reconstruction consistent with the available measurements and with the implicit regularization induced by the neural parametrization.

\section{Methodology}
\label{sec 3: methodology}

In this section, we introduce the methodology that will be used for reconstructing the underlying electric conductivity $\gamma(\mathbf{x})$ in the domain $\Omega$ (the unit square in our case) from a finite set of boundary measurements of potential $f$ and induced current $J$. Our approach leverages PINNs as a tool to invert this problem. We first introduce PINNs in a general setting, and review several of their applications. Second, we explain the generation of synthetic data via numerical integration of the direct problem. Finally, we set up the PINNs pipeline to tackle the inverse problem.

\subsection{Physics-Informed Neural Networks} \label{subsec: intro to pinns}

PINNs were originally introduced by Dissanayake and Phan-Thien~\cite{Dissanayake} and Lagaris et al.~\cite{lagaris_orig_PINNs} as a framework for solving ordinary and partial differential equations (ODEs/PDEs) using neural networks (NNs) as function approximators, and were later significantly advanced by Raissi et al.~\cite{RAISSI2019686}, who demonstrated their effectiveness across a wide range of challenging physical systems. More broadly, the use of NNs for differential equations predates PINNs and includes notable approaches such as the Deep Galerkin Method~\cite{Sirignano2018DGM}, physics-constrained learning techniques~\cite{Zhu2019PhysicsConstrained}, and symmetry-aware architectures~\cite{protopapas_sondiak_Syms_in_Pinns}. Within the PINNs framework, each unknown function is typically approximated by a separate NN, most commonly a multilayer perceptron (MLP), which consists of stacked affine transformations interleaved with nonlinear activation functions. These models are characterized by a large number of trainable parameters, known as weights and biases, which are optimized during the training process by minimizing a loss function. In PINNs, this loss function is constructed from the squared residuals of the governing differential equations, together with contributions from boundary and/or initial conditions (BCs/ICs), thereby embedding the underlying physical laws directly into the learning objective. Further developments in this direction aim to incorporate additional physical structure into NNs, for instance by enforcing conservation laws or Hamiltonian dynamics, as explored in~\cite{Choudhary2020PhysicsEnhanced, Choudhary2020Forecasting, Greydanus2019HNN}.

Despite the significant progress achieved in recent years, PINNs are still generally outperformed by classical numerical methods in terms of accuracy and computational efficiency, particularly for high-dimensional or stiff problems~\cite{karniadakis2021physics}. Nevertheless, PINNs offer a number of distinctive advantages. First, they are inherently mesh-free, meaning that the learned solution is not tied to a predefined discretization grid. Instead, it can be evaluated at arbitrary points within the domain, enabling flexible and efficient interpolation across the training region. Second, PINNs can be extended to learn parametric families of solutions, often referred to as ``bundle solutions''~\cite{flamant2020solving}. In this setup, the network is trained to capture the dependence of the solution on varying parameters, BCs, or even different forms of the governing equations. In contrast to traditional solvers, which typically require re-running the full numerical pipeline for each new configuration, this approach allows a single trained model to provide rapid predictions across a range of scenarios. Finally, and most relevant for this work, PINNs are particularly well-suited for inverse problems~\cite{RAISSI2019686, yang2021bayesian, Bea:2024xgv,tarancon2025efficient}, where the goal is to infer unknown functions or parameters appearing in the differential equations from indirect observations, such as boundary measurements. This capability stems from their ability to efficiently evaluate solutions under varying inputs, combined with the flexibility of the learning framework to incorporate observational data directly into the training process.

The Calder\'on problem represents a compelling test case for PINN-based inversion. It is governed by a well-understood elliptic PDE, yet recovering the interior conductivity from boundary data alone is highly unstable, since small perturbations in the measurements can lead to substantial changes in the reconstruction. This makes it an ideal setting in which to assess both the flexibility and the limitations of the PINN framework. The approach developed in this work builds directly on the capabilities described above: the mesh-free nature of PINNs allows the conductivity and potential fields to be represented continuously across the domain, while the inverse problem formulation enables both unknown fields to be recovered simultaneously from a finite collection of boundary measurements. In contrast to methods that require large training datasets or supervised pretraining, our framework treats each reconstruction as a self-contained optimization problem, driven entirely by the PDE residual and the available DtN data. The following sections describe the specific architecture, training procedure, and boundary excitation strategy that constitute our proposed methodology.

\subsection{Data generation}\label{subsec: data generation}

We consider a fully synthetic setting in which a ground-truth conductivity
$\gamma^{\mathrm{true}}$ is prescribed and the corresponding boundary
measurements $\{f_k,J_k\}_{k=1}^{K}$ are generated by solving the forward
problem numerically. For each imposed Dirichlet condition $f_k$, the
potential $u_k$ is computed on a uniform $N\times N$ grid using a
finite-difference discretization of the divergence-form operator with
harmonic averaging of interface conductivities. The resulting sparse
linear system is solved using \texttt{scipy.sparse.linalg.spsolve}, and the
boundary current $J_k=\gamma\partial_nu_k$ is evaluated using one-sided
finite differences. Further details are provided in~\ref{appendix:fdm solver}.

A crucial aspect of the Calder\'on problem is the choice of BCs $f_k$ used to probe the interior of the domain. Since the conductivity $\gamma(x)$ is recovered from the reduced DtN map $\Lambda_\gamma(f)$, the quality of the reconstruction strongly depends on the chosen boundary excitations. While uniqueness is guaranteed when the full DtN operator
\begin{equation}
\Lambda_\gamma : H^{1/2}(\partial\Omega) \rightarrow H^{-1/2}(\partial\Omega)
\end{equation}
is known for all admissible boundary potentials $f \in H^{1/2}(\partial\Omega)$, in practice only a finite number of BCs can be imposed. Therefore, the selection of informative BCs becomes crucial for accurately reconstructing $\gamma(x)$.

We use randomized localized wavelet boundary excitations in order to probe the domain at multiple locations and spatial scales. This choice was motivated by preliminary empirical comparisons with trigonometric, Gaussian, and linear boundary conditions, in which the wavelet-based excitations consistently produced more accurate conductivity reconstructions. We stress that we are solving the problem on a unit square, so $\Omega = [0,1] \times [0,1]$. On each active side of the square boundary,
we prescribe a superposition of tapered Ricker wavelets,
\begin{equation}
    f(x)=\sum_{i=1}^{N_w}
    A_i\,\psi(x;x_i,a_i)\sin(\pi x),
    \qquad x\in[0,1],
\end{equation}
where
\begin{equation}
    \psi(x;x_i,a_i)=
    \left[1-2\bigl(\pi a_i(x-x_i)\bigr)^2\right]
    \exp\left[-\bigl(\pi a_i(x-x_i)\bigr)^2\right].
\end{equation}
Here, \(x_i\) is the center of the \(i\)-th wavelet, \(A_i\) is its amplitude, \(a_i\) controls its spatial scale, and \(N_w\) is the number of wavelets placed on the side. The zeros surrounding the positive central lobe occur at
\begin{equation}
    x=x_i\pm\frac{1}{\sqrt{2}\pi a_i},
\end{equation}
so that its characteristic zero-to-zero width is
\begin{equation}
    w_i=\frac{\sqrt{2}}{\pi a_i}.
\end{equation}
Thus, larger values of \(a_i\) correspond to more localized boundary excitations. For the range \(a_i\in[0.3,3]\), the central-lobe width varies approximately from \(0.15\) to \(1.50\) in units of the side length. Since the Ricker wavelet has exponentially decaying tails and is multiplied by the tapering factor \(\sin(\pi x)\), this width should be understood as a characteristic scale rather than a compact support. The taper ensures that the imposed potential vanishes at the corners.

For each BC, the active sides, number of wavelets, centers, scales, and amplitudes are sampled randomly. We use
\begin{equation}
    x_i\sim\mathcal{U}([0.2,0.8]),
    \qquad
    a_i\sim\mathcal{U}([0.3,3]),
    \qquad
    A_i\sim\mathcal{U}([0.5,4.5]).
\end{equation}
Up to three wavelets are superposed on each active side, and the number of active sides is sampled from \(\{1,2,3,4\}\).

\begin{figure}[t]
    \centering
    \includegraphics[width=\textwidth,height=0.5\textheight,keepaspectratio]{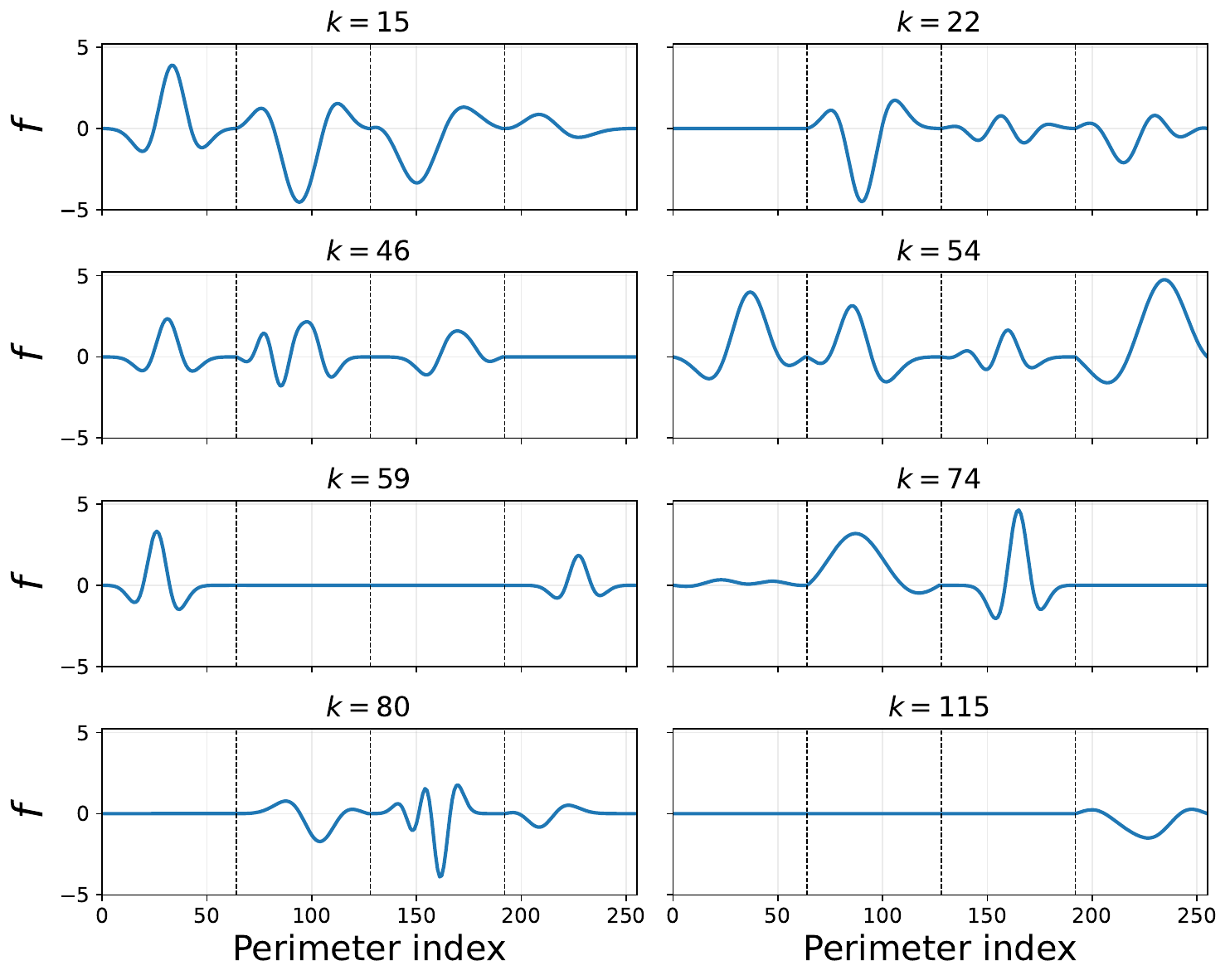}
    \caption{Examples of wavelet-based BCs used in the data generation process. $k$ represents the index labeling different BC, running from $1$ to $K$. The perimeter index traverses the boundary counterclockwise, starting at the bottom-left corner, in the order bottom, right, top, and left.} 
    \label{fig:wavelet_bcs}
\end{figure}

The use of superpositions provides boundary excitations containing several localized scales and positions within a single measurement. This is a practical compromise: the space of admissible boundary functions is infinite-dimensional, so it is impossible to probe every location and spatial scale. Moreover, each additional BC requires a separate forward solve and increases the amount of data and computational cost of PINN training. Likewise, using arbitrarily many or arbitrarily narrow wavelets would eventually exceed the resolution of the finite boundary grid. The randomized finite superpositions therefore provide a computationally manageable approximation to a richer multiscale family of boundary probes, without claiming to constitute an optimal set of excitations. Examples are shown in Fig.~\ref{fig:wavelet_bcs}.

\subsection{Inverse problem with PINNs} \label{subsec: PINNs Approach}

Given the finite boundary dataset
$\{f_k,J_k\}_{k=1}^{K}$, we seek a conductivity field
$\gamma^{\mathrm{NN}}$ and corresponding potential fields
$\{u_k^{\mathrm{NN}}\}_{k=1}^{K}$ that satisfy the governing PDE and
reproduce the measured boundary responses, i.e. Eq. \eqref{eq: calderon ODE + BC}.

\begin{figure}[t]
    \includegraphics[trim=30 30 0 30, clip, width=\textwidth]{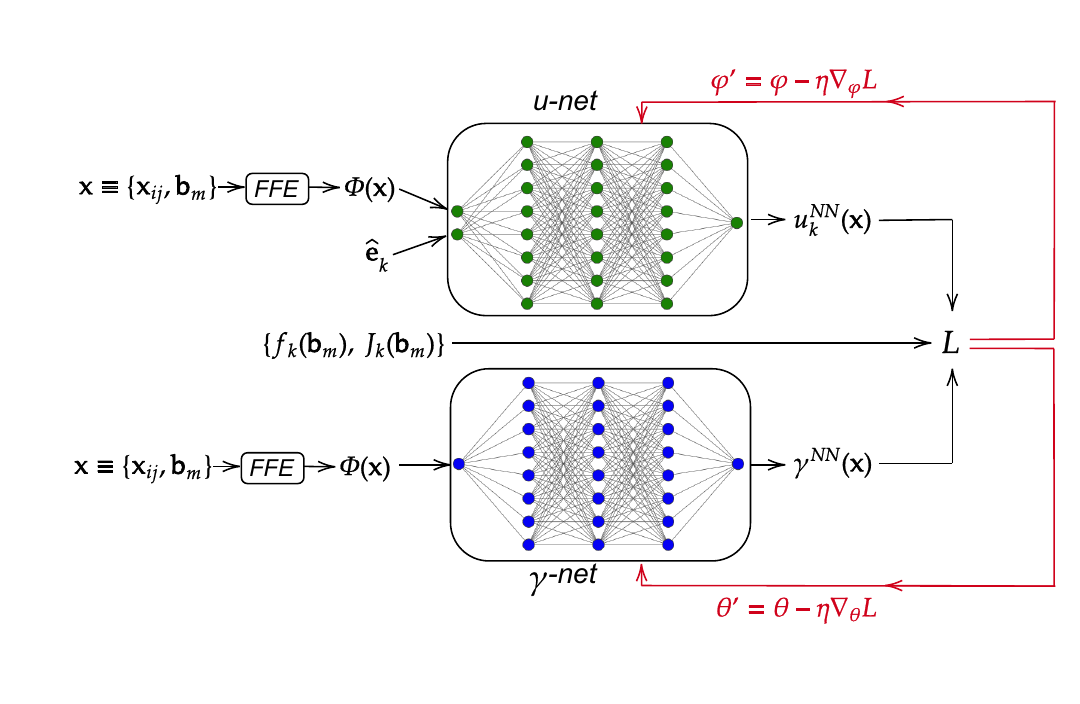}
    \caption{PINN architecture used for conductivity reconstruction. The
potential network \textit{u-net} takes the spatial coordinates and a one-hot
boundary-condition label $\mathbf e_k$ and predicts
$u_k^{\mathrm{NN}}$. The conductivity network \textit{$\gamma$-net} predicts the
shared field $\gamma^{\mathrm{NN}}$. Spatial coordinates are either used
directly or mapped through the Fourier feature encoder $\Phi$. The two
networks are optimized jointly using the PDE, Dirichlet, and Neumann losses
defined in Eq.~\eqref{eq: loss tot}. A forward pass is represented by black arrows, while backpropagation and updating of the NNs' parameters is represented by red arrows.}
    \label{fig: pinns setup}
\end{figure}

We adopt a two NN architecture depicted in Fig. \ref{fig: pinns setup}. The first NN, denoted from now on as \textit{u-net}, takes as inputs two coordinates $(x,y)$ and an encoding \textit{one-hot vector} $\mathbf{e}_k$, and gives as output $K$ functions acting on the 2D domain, denoted by $u_k^\text{NN}(x,y)$, where $k=1,\dots,K$. The second NN, denoted as $\gamma$-\textit{net}, takes as input two coordinates $(x,y)$ and gives as output a single function $\gamma^\text{NN}(x,y)$ acting on the 2D domain. The overall pipeline is summarized step-by-step below:

\begin{enumerate}
    \item A discrete sampling (grid) of points $(x_i,y_j)\equiv\mathbf{x}_{ij}$ of the 2D interior of the domain $\text{int}(\Omega) \equiv \Omega \backslash \partial \Omega=~(0,1)\times(0,1)$ are given as inputs to both NNs (the u-net and the $\gamma$-net), where $i,j=1,\dots,N$. 
    The sampled boundary points are denoted by
    \[
    \mathbf{b}_m=(x_m,y_m)\in\partial\Omega,
    \qquad m=1,\dots,4N_b-4,
    \]
    where the index \(m\) follows an ordered traversal of the square boundary
    (bottom \(\rightarrow\) right \(\rightarrow\) top \(\rightarrow\) left). $N_b-1$ is the number of points on each side.
    \item A set of one-hot encoding vectors $\mathbf{e}_k$ ($k=1,\dots,K)$ are given as input to the u-net. Each of these vectors encodes the $k$-th boundary data pair that the u-net is solving for a particular instance during the training. The only purpose of this vector is to select the $k-$th initial condition of the u-net.

    \item The u-net outputs $K$ different functions $u_k^\text{NN}$ evaluated on the grid, both on the interior points and the boundary points.
    
    \item The $\gamma$-net outputs a \textbf{single} function $\gamma^\text{NN}$ evaluated on the grid,  both on the interior points and the boundary points.
    
    \item The loss function is defined through the $L2$-norm of the residuals of the DE and both types of BCs:

\begin{equation}
    \label{eq: loss tot}
    L = \lambda_\text{DE} L_\text{DE} + \lambda_\text{f} L_\text{Dirichlet} + \lambda_\text{J} L_\text{Neumann} \ \ ,
\end{equation}
where:

\begin{align}
    & L_\text{DE} = \sum_{i,j=1}^N\sum_{k=1}^K \left[\nabla\cdot\left(\gamma^\text{NN}(\mathbf{x}_{ij})\nabla u_k^\text{NN}(\mathbf{x}_{ij})\right)\right]^2 \ \ ,  \label{eq: loss DE}  \\
     & L_\text{Dirichlet} = \sum_{m=1}^{4N_{b}-4} \sum_{k=1}^K \left[u^\text{NN}_k(\mathbf{b_m}) - f_k(\mathbf{b_m})\right]^2  \, , \label{eq: dirichlet loss BC} \\
     & L_\text{Neumann} = \sum_{m=1}^{4N_{b}-4} \sum_{k=1}^K \left[\gamma^\text{NN}(\mathbf{x})\left.\frac{\partial u^\text{NN}_k(\mathbf{x})}{\partial \hat{n}}\right|_{\mathbf{x}=\mathbf{b_m}} - J_k(\mathbf{b_m})\right]^2 \,, \label{eq: neumann loss BC}
\end{align}

being $\lambda_\text{DE}$, $\lambda_\text{f}$ and $\lambda_\text{J}$ tunable relative weights (hyperparameters of the model). These values are chosen to be $\lambda_\text{DE} = 1, \ \lambda_\text{f} = \{10,\,10^3\}, \ \lambda_\text{J}=10$, based on phenomenological observation of the convergence of the loss function terms. For the exact details, see Table \ref{tab:training_losses}.

    \item Backpropagation and updating of both NNs' parameters is performed through stochastic gradient descent optimization (\textit{Adam} optimizer \cite{AdamOptimizerkingma2017}). The decrease of the total loss makes the outputs of the \textit{u-net} converge to approximate solutions for each of the BCs, and the output of $\gamma$\textit{-net} converges to an approximate inversion of the spatially-dependent electric conductivity, as described in detail in the text below.
\end{enumerate}

The PDE term \eqref{eq: loss DE} enforces the conductivity equation at interior collocation
points, while the Dirichlet term \eqref{eq: dirichlet loss BC} and Neumann term \eqref{eq: neumann loss BC} penalize discrepancies with the measured boundary potentials and currents. These boundary constraints are imposed softly\footnote{The imposition of the BCs is a \textit{soft} enforcing, meaning that the BCs are satisfied up to the error in the BC loss function, as opposed to a \textit{hard} enforcing, where the NN output is reparametrized to satisfy the BCs exactly, see e.g. \cite{lagaris_orig_PINNs, protopapas_sondiak_Syms_in_Pinns, lagaris_irreg_boundaries}.} through the loss function. Both networks are optimized
jointly by backpropagation using the \textit{Adam} optimizer.

For further specifications on the technical aspects and the hyperparameter choices made in the training phase, see \ref{appendix: training and loss specifications}.

\subsubsection{Fourier Feature Encoding} \label{subsec: ffe}

A central architectural choice in this work is the use of Fourier Feature Encoding (FFE) for the spatial coordinates. This choice is motivated by the fact that many physically relevant conductivity profiles in EIT are not globally smooth. In particular, inclusions, layered media, and sharp material interfaces naturally introduce localized or discontinuous structures in $\gamma(x,y)$. Standard coordinate-based multilayer perceptrons (MLPs), however, are known to exhibit a \emph{spectral bias}: during training, they tend to learn low-frequency components of a target function before high-frequency ones, and they often represent sharp transitions only after long training times or with significant smoothing \cite{spectral_bias_Rahaman_2019}. This effect is especially relevant for inverse problems such as the Calder\'on problem, where the available data are already indirect and boundary-limited.

Previous PINN-based approaches to related EIT settings often considered smooth conductivity phantoms, for example by applying Gaussian smoothing to the target profiles, and used raw spatial coordinates as network inputs \cite{Pokkunuru2023,BarSochen2021}. In contrast, one of the goals of the present work is to test whether the PINN framework can recover less regular conductivities. For this reason, we include FFE as a way of enriching the coordinate representation before passing it to the NNs. The purpose is not to impose a particular Fourier-series ansatz on the conductivity, but rather to provide the MLP with input features oscillating at multiple spatial frequencies, making high-frequency or localized variations easier to represent.

Concretely, before being passed to the NNs, each spatial coordinate
\begin{equation}
\mathbf{x}=(x,y)\in\Omega
\end{equation}
is mapped to a higher-dimensional feature vector $\Phi(\mathbf{x})$ with
\begin{equation}
\Phi:\mathbb{R}^2\rightarrow\mathbb{R}^{2M}
\end{equation}
that populates the input space with Fourier frequencies given by a fixed matrix \(W\in\mathbb{R}^{M\times 2}\) whose entries are sampled from a random Gaussian distribution. The integer \(M\) controls the number of sampled Fourier modes, so that the encoded coordinate has dimension \(2M\).
Mainly,
\begin{equation}
W =
\begin{pmatrix}
\boldsymbol{\omega}_1^T \\
\boldsymbol{\omega}_2^T \\
\vdots \\
\boldsymbol{\omega}_M^T
\end{pmatrix}
\in \mathbb{R}^{M\times 2},
\qquad 
\boldsymbol{\omega}_m=(\omega_{m1},\omega_{m2}),
\label{eq:ffe_matrix}
\end{equation}
and define the Fourier-feature map as

\begin{equation}
\Phi(\mathbf{x}) =
\begin{pmatrix}
\sin(2\pi \boldsymbol{\omega}_1\cdot \mathbf{x}) \\
\vdots \\
\sin(2\pi \boldsymbol{\omega}_M\cdot \mathbf{x}) \\
\cos(2\pi \boldsymbol{\omega}_1\cdot \mathbf{x}) \\
\vdots \\
\cos(2\pi \boldsymbol{\omega}_M\cdot \mathbf{x})
\end{pmatrix} = \begin{pmatrix}
\sin\!\left(2\pi(\omega_{11}x+\omega_{12}y)\right) \\
\vdots \\
\sin\!\left(2\pi(\omega_{M1}x+\omega_{M2}y)\right) \\
\cos\!\left(2\pi(\omega_{11}x+\omega_{12}y)\right) \\
\vdots \\
\cos\!\left(2\pi(\omega_{M1}x+\omega_{M2}y)\right)
\end{pmatrix}
\in \mathbb{R}^{2M}.
\label{eq:ffe_vector}
\end{equation}

The scale of the random Gaussian distribution used to sample \(W\) controls the typical frequencies present in the encoding: small values emphasize slowly varying features, while larger values make higher-frequency spatial variations accessible to the NNs.

In the implementation used throughout this work, the Fourier matrix has
\(M=256\) rows, and its entries are sampled independently according to
\begin{equation}
    W_{ij}=\sigma_W Z_{ij},
    \qquad
    Z_{ij}\sim\mathcal{N}(0,1),
    \qquad
    \sigma_W=10.
\end{equation}
Equivalently,
\begin{equation}
    W_{ij}\overset{\mathrm{i.i.d.}}{\sim}
    \mathcal{N}(0,\sigma_W^2)
    =
    \mathcal{N}(0,100),
\end{equation}
where the second argument denotes the variance. Thus, \(W\in
\mathbb{R}^{256\times 2}\), and the concatenation of the sine and cosine
components produces an encoded coordinate vector
\(\Phi(\mathbf{x})\in\mathbb{R}^{512}\). The matrix \(W\) is sampled once
at initialization and remains fixed and non-trainable throughout the
optimization. 

This encoding is applied to the input coordinates of both the conductivity network and the potential network. Thus, instead of learning $\gamma^{\rm NN}(x,y)$
directly from the raw coordinates, the conductivity network learns the composition
\begin{equation}
\gamma^{\rm NN}(\mathbf{x}) =
G_\theta\!\left(\Phi(\mathbf{x})\right),
\label{eq:gamma_net_ffe}
\end{equation}
where \(G_\theta\) denotes the MLP representing the conductivity, being $\theta$ the weights and biases of the $\gamma$-net. 
Similarly, the potential network learns
\begin{equation}
u_k^{\rm NN}(\mathbf{x}) =
U_\varphi\!\left(\Phi(\mathbf{x}),\mathbf e_k\right),
\label{eq:u_net_ffe}
\end{equation}
where $\varphi$ are the weighs and biases of the $u$-net. In this way, the same coordinate transformation is used consistently in both networks entering the PDE residual\footnote{We note that the derivatives present in the Calderon equation \eqref{eq:pde_residual_ffe} are always taken with respect to the raw-coordinates $\mathbf{x}$, and that the chain-rule implicit in the compositions $G_\theta(\Phi(\mathbf{x}))$, and $U_\varphi(\Phi(\mathbf{x}))$ is computed through automatic differentiation.}
\begin{equation}
\nabla\cdot\left(\gamma^{\rm NN}\nabla u_k^{\rm NN}\right).
\label{eq:pde_residual_ffe}
\end{equation}

The main motivation for using FFE is therefore to modify the low-frequency bias of standard coordinate-based MLPs by making higher-frequency spatial variations more accessible to the network. Although the resulting neural representations remain continuous and smooth, the Fourier-encoded inputs allow the networks to approximate sharper transitions and more localized features than may be readily captured from raw coordinates alone. In the inverse Calder\'on problem, sharp inclusions or interfaces can generate boundary responses that are subtle and difficult to distinguish from smooth alternatives, especially when only finitely many boundary measurements are available. A raw-coordinate MLP may fit the boundary data while still representing the interior conductivity with a smoothed approximation. By enriching the input space with sinusoidal features, the network is better able to represent localized transitions and higher-frequency spatial content, which can improve the recovery of inclusions and discontinuous profiles.

FFE should be understood as an architectural tool that changes the implicit bias of the neural representation. In the results section, we therefore compare reconstructions obtained with and without FFE in order to assess when this enriched representation improves the practical recovery of high-frequency or discontinuous conductivity structures.

\subsection{Metrics used in evaluation} 
\label{subsec:evaluation_metrics}

We evaluate the reconstructed conductivity \(\gamma^{\rm NN}(\mathbf{x})\) using a set of complementary metrics. The reason for using several metrics is that the quality of an inverse reconstruction is not captured by a single scalar number. Pointwise error measures amplitude accuracy, structural metrics measure similarity of spatial patterns, geometric metrics quantify the recovery of localized regions, and correlation metrics assess whether the global spatial organization of the conductivity has been captured.

Pointwise accuracy is measured using the relative error (RE)
\begin{equation}
    RE(\mathbf{x})=
    \frac{|\gamma^{\rm NN}(\mathbf{x})-\gamma^{\rm true}(\mathbf{x})|}
    {|\gamma^{\rm true}(\mathbf{x})|},
    \label{eq:relative_error_gamma}
\end{equation}
together with the mean squared error (MSE). These metrics are most useful when the absolute amplitude of the recovered conductivity is important, for example in smooth profiles where the reconstruction error is distributed over the whole domain. However, pointwise errors can be less informative for inclusion-type profiles, since a small spatial shift of an otherwise correctly reconstructed inclusion may lead to a large MSE.

To assess structural agreement, we compute the Structural Similarity Index Measure (SSIM), originally introduced in \cite{SSIM_orig} and also used for reconstruction quality assessment in \cite{BarSochen2021,Yang2024CPFI}. SSIM compares local luminance, contrast, and structural information between \(\gamma^{\rm NN}\) and \(\gamma^{\rm true}\). In our setting, it is useful for determining whether the reconstructed conductivity preserves the qualitative spatial structure of the target profile, especially for smooth or heterogeneous conductivities. Unlike MSE, SSIM is less purely pointwise and is therefore more sensitive to the preservation of patterns and contrast. 

For inclusion-type problems, where the main goal is often to recover the position, shape, and extent of anomalous regions, we use the Intersection over Union (IoU). Given a threshold \(\Gamma\), we define binary masks
\begin{equation}
    A^{\rm label}(\mathbf{x}) =
    \begin{cases}
         1, & \gamma^{\rm label}(\mathbf{x})\geq \Gamma,\\
         0, & \gamma^{\rm label}(\mathbf{x})< \Gamma,
    \end{cases}
    \qquad
    {\rm label}\in\{{\rm NN},{\rm true}\},
\end{equation}
and compute
\begin{equation}
    \mathrm{IoU} =
    \frac{|A^{\rm NN}\cap A^{\rm true}|}
    {|A^{\rm NN}\cup A^{\rm true}|}.
    \label{eq:iou_metric}
\end{equation}
This metric is particularly well suited for sharp inclusions and piecewise-constant profiles, because it directly measures whether the reconstructed high-conductivity or low-conductivity region overlaps with the true one. It is less appropriate as a primary metric for smooth profiles, where the choice of threshold is less physically natural. The threshold $\Gamma$ is chosen to be at the mean value of the true conductivity, $\Gamma = \Bar{\gamma}^\text{true}$.

We also compute the Pearson correlation coefficient
\begin{equation}
    r_{xy}= 
    \frac{\sum_{i=1}^N (x_i-\mu_x)(y_i-\mu_y)}
    {\sqrt{\sum_{i=1}^N (x_i-\mu_x)^2}
    \sqrt{\sum_{i=1}^N (y_i-\mu_y)^2}},
    \label{eq:pearson_correlation_formula}
\end{equation}
where \(x_i\) and \(y_i\) denote the values of \(\gamma^{\rm true}\) and \(\gamma^{\rm NN}\) at the \(i\)-th spatial point, respectively. $\mu_x,\,\mu_y$ are the corresponding mean values. Pearson correlation measures whether high and low conductivity regions appear in the correct locations, independently of an overall affine rescaling. It is therefore useful for random or heterogeneous profiles, where recovering the global spatial organization may be more important than matching every amplitude exactly.

For completeness, we also report the peak signal-to-noise ratio (PSNR), which has been used in related reconstruction studies \cite{BarSochen2021,Pokkunuru2023}. PSNR is derived from the MSE and therefore primarily reflects pointwise amplitude accuracy. Larger values indicate smaller reconstruction error.

Finally, we also monitor the recovery of the boundary data. Given the measured DtN pairs \(\{f_k,J_k\}\), the PINN predicts corresponding quantities \(\{f_k^{\rm NN},J_k^{\rm NN}\}\). The discrepancy between these pairs is precisely controlled by the Dirichlet and Neumann loss terms in Eqs.~\eqref{eq: dirichlet loss BC} and~\eqref{eq: neumann loss BC}. These boundary losses measure how well the learned fields reproduce the available finite samples of the DtN map.

\section{Results and Discussion}
\label{sec: results}

We present the reconstruction results obtained with the proposed PINN framework, with and without an FFE reparameterization of the input space.
The experiments are divided into three groups. First, we consider discontinuous or sharp
conductivity profiles, where the effect of FFE is most visible.
Second, we consider smooth profiles (with low Fourier frequencies), for which the raw-coordinates representations are expected to perform competitively. Third, we consider profiles generated with frequencies sampled from a random Gaussian distribution, that present an intermediate behavior between the two previous groups. Additionally, we analyze the frequency distribution of a representative example for each group and how it connects to the usefulness of FFE, and analyze the sensitivity of the reconstruction method to the central regions of the domain, where instability is stronger. For the explicit formulas and parameter values defining each conductivity profile, see \ref{ap:conductivity_profiles}. For training, architecture, and loss specifications and values, see \ref{appendix: training and loss specifications}.

\subsection{Sharp and inclusion-type conductivities}

\begin{figure}[htbp]
    \centering
    \includegraphics[width=\textwidth, keepaspectratio]{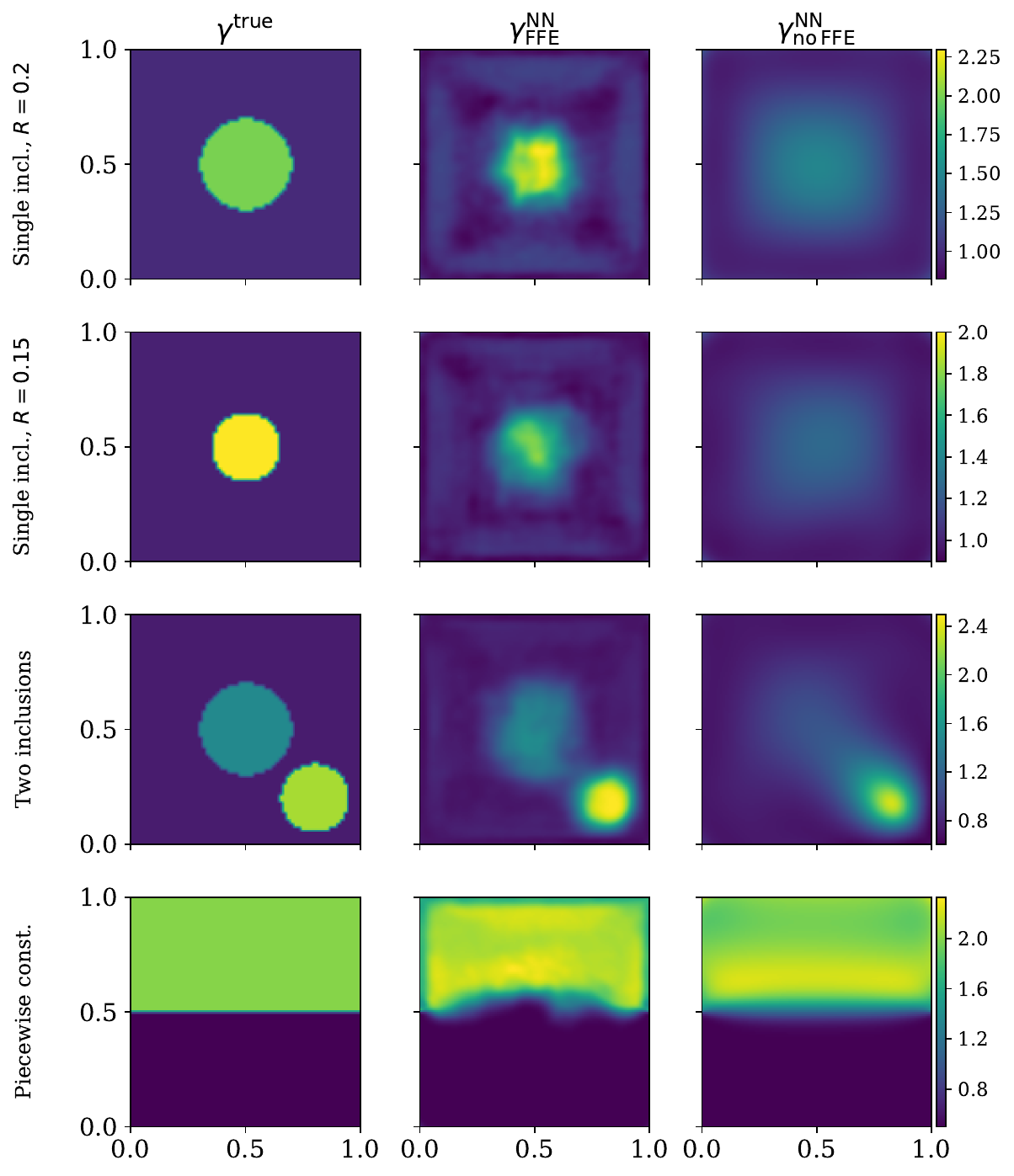}
    \caption{Reconstructed conductivities for sharp, discontinuous profiles. We compare the ground through (left column) with PINN reconstructions with FFE (middle column) and without FFE (right column). Each row corresponds to a different ground-truth conductivity. The conductivity/experiment label is indicated to the left side of each row.}
    \label{fig:inclusions_sharp_reconstructions}
\end{figure}

Reconstructions obtained with and
without FFE for three sharp conductivity profiles are compared in Figure~\ref{fig:inclusions_sharp_reconstructions}. Each row
corresponds to a different ground-truth conductivity profile/different experiment, while the columns show the true
conductivity, the reconstruction obtained using FFE, and the reconstruction
obtained using raw coordinates (without FFE). 


\begin{table}[htbp]
\centering
\resizebox{\textwidth}{!}{
\begin{tabular}{llcccccc}
\hline
\multicolumn{1}{c}{\textbf{Experiment}} &
& \multicolumn{1}{c}{\textbf{RE} $\downarrow$} 
& \multicolumn{1}{c}{\textbf{MSE} $\downarrow$} 
& \multicolumn{1}{c}{\textbf{SSIM} $\uparrow$} 
& \multicolumn{1}{c}{\textbf{IoU} $\uparrow$} 
& \multicolumn{1}{c}{\textbf{Pearson} $\uparrow$} 
& \multicolumn{1}{c}{\textbf{PSNR}} $\uparrow$ \\ 
\hline

\multirow{2}{*}{Single Inclusion $R=0.2$}
& FFE    
& 0.080 & 0.027 & 0.775 & 0.799 & 0.866 & 15.67 \\
& no FFE 
& 0.106 & 0.053 & 0.896 & 0.388 & 0.767 & 12.77 \\
\hline

\multirow{2}{*}{Single Inclusion $R=0.15$}
& FFE    
& 0.060 & 0.019 &	0.868 & 0.433 & 0.872 & 17.33\\
& no FFE 
& 0.080	& 0.043	& 0.939	& 0.215 &	0.601	& 13.65 \\
\hline

\multirow{2}{*}{Two Inclusions}
& FFE    
& 0.088 & 0.036 & 0.843 & 0.802 & 0.904 & 18.00 \\
& no FFE 
& 0.134 & 0.063 & 0.859 & 0.726 & 0.840 & 15.55 \\
\hline

\multirow{2}{*}{Piecewise Constant}
& FFE    
& 0.072 & 0.058 & 0.847 & 0.934 & 0.949 & 15.86 \\
& no FFE 
& 0.059 & 0.022 & 0.900 & 0.979 & 0.980 & 20.02 \\
\hline

\end{tabular}}
\caption{Comparison of reconstruction metrics between FFE and no-FFE models for the first set of conductivity profiles. Upward arrows indicate that larger values of the metric correspond to better reconstructed profiles, while downward arrows indicate the opposite.}
\label{tab:metrics_table_1}
\end{table}

\begin{figure}[t]
    \centering

    \begin{subfigure}{0.49\textwidth}
        \centering
        \includegraphics[
            trim = 0 0 1040 25,
            clip,
            width=\textwidth
        ]{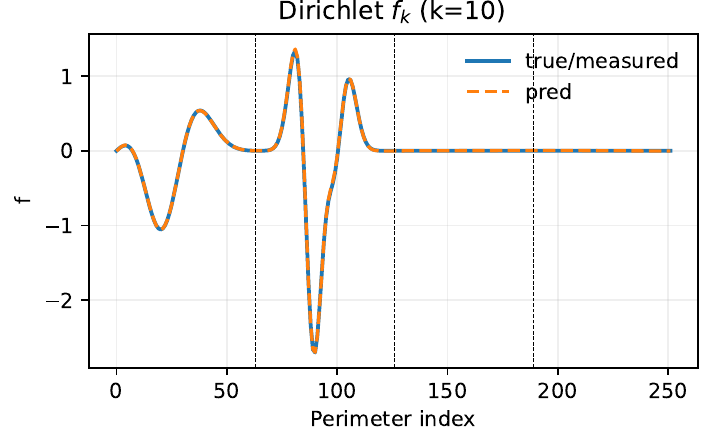}
        \caption{}
        \label{fig:boundary_fit_a}
    \end{subfigure}
    \hfill
    \begin{subfigure}{0.49\textwidth}
        \centering
        \includegraphics[
            trim = 350 0 692 25,
            clip,
            width=\textwidth
        ]{Plots/boundary_fit_k_10_epoch_291000.pdf}
        \caption{}
        \label{fig:boundary_fit_b}
    \end{subfigure}

    \vspace{0.5em}

    \begin{subfigure}{0.49\textwidth}
        \centering
        \includegraphics[
            trim = 682 0 357 25,
            clip,
            width=\textwidth
        ]{Plots/boundary_fit_k_10_epoch_291000.pdf}
        \caption{}
        \label{fig:boundary_fit_c}
    \end{subfigure}
    \hfill
    \begin{subfigure}{0.49\textwidth}
        \centering
        \includegraphics[
            trim = 1040 0 0 25,
            clip,
            width=\textwidth
        ]{Plots/boundary_fit_k_10_epoch_291000.pdf}
        \caption{}
        \label{fig:boundary_fit_d}
    \end{subfigure}

    \caption{
    Comparison between the true data and PINN predictions for a representative BC (out of the set of BCs) for the single inclusion $R=0.2$ profile.
    \textbf{(a)}: Dirichlet BC $f$ (or boundary potential) data (solid blue) and PINN prediction (dashed orange). \textbf{(b)}: Residuals for the Dirichlet BC prediction. \textbf{(c)}: Neumann BC $J$ (or boundary current) data (solid blue) and PINN prediction (dashed orange). \textbf{(d)}: Residuals for the Neumann BC prediction.
    }
    \label{fig:representative_boundary_fit}
\end{figure}

Across the inclusion-type examples, FFE generally improves the recovery of localized structures. For the two single-inclusion cases and the two-inclusion case, the FFE model gives lower relative error and MSE, higher IoU, higher Pearson correlation, and higher PSNR than the corresponding raw-coordinate model. The improvement in IoU is particularly relevant here, since these profiles are defined by sharp anomalous regions and the main goal is to recover their location and geometry. For example, in the single-inclusion case with \(R=0.2\), the IoU increases from \(0.388\) without FFE to \(0.799\) with FFE, indicating a substantially better overlap between the recovered and true inclusion.

Interestingly, the no-FFE model gives larger SSIM values in the single-inclusion cases, despite performing worse in IoU and point-wise error. This illustrates that SSIM can favor globally smooth structural similarity and does not always reflect the accuracy of sharp interface localization. For inclusion-type profiles, IoU and point-wise errors are therefore more informative diagnostics than SSIM alone.

The piecewise-constant horizontal split is an exception to the general trend: in this case, the raw-coordinate model performs better across all reported metrics. This suggests that the advantage of FFE is not universal, but depends on the geometry of the target conductivity and particularly on the geometry of the discontinuity. For this particular case, the discontinuity happens across the $y$-direction exclusively, while the $x$-direction is constant for all $y$-values. Therefore, the high-frequency modes are only non-vanishing in the $y$-direction, while the $x$-direction can be represented by a single $0$-frequency mode. FFE populates the input space with high frequencies approximately homogeneously and isotropically, therefore for this particular case, the NNs need to suppress all high-frequency Fourier modes modes in the $x$-direction for all values of $y$. This makes the training for this particular case more contrived for the FFE method, even if a discontinuity is present.

Furthermore, we show only one representative boundary fit in Fig.~\ref{fig:representative_boundary_fit}. The figure compares the imposed (theoretical) DtN data pair $(f_k,J_k)$ with the corresponding PINN predictions, for one BC in the single-inclusion case with radius 0.2 with FFE. Similar agreement is obtained across the other BCs and conductivity profiles, with and without FFE. We observe a fit with residual around or less than $1\%$ in the Dirichlet BC $f$ and around or less than $0.1\%$ in the Neumann BC $J$. This indicates that the obtained reconstruction for the conductivity $\gamma^\text{NN}(\mathbf{x})$ is compatible with the boundary DtN data pairs up to the error given by the loss function value achieved (see Table \ref{tab:training_losses} for the specific values).

\subsection{Smooth conductivities}

\begin{figure}[htbp]
    \centering
    \includegraphics[width=0.7\textwidth, keepaspectratio]{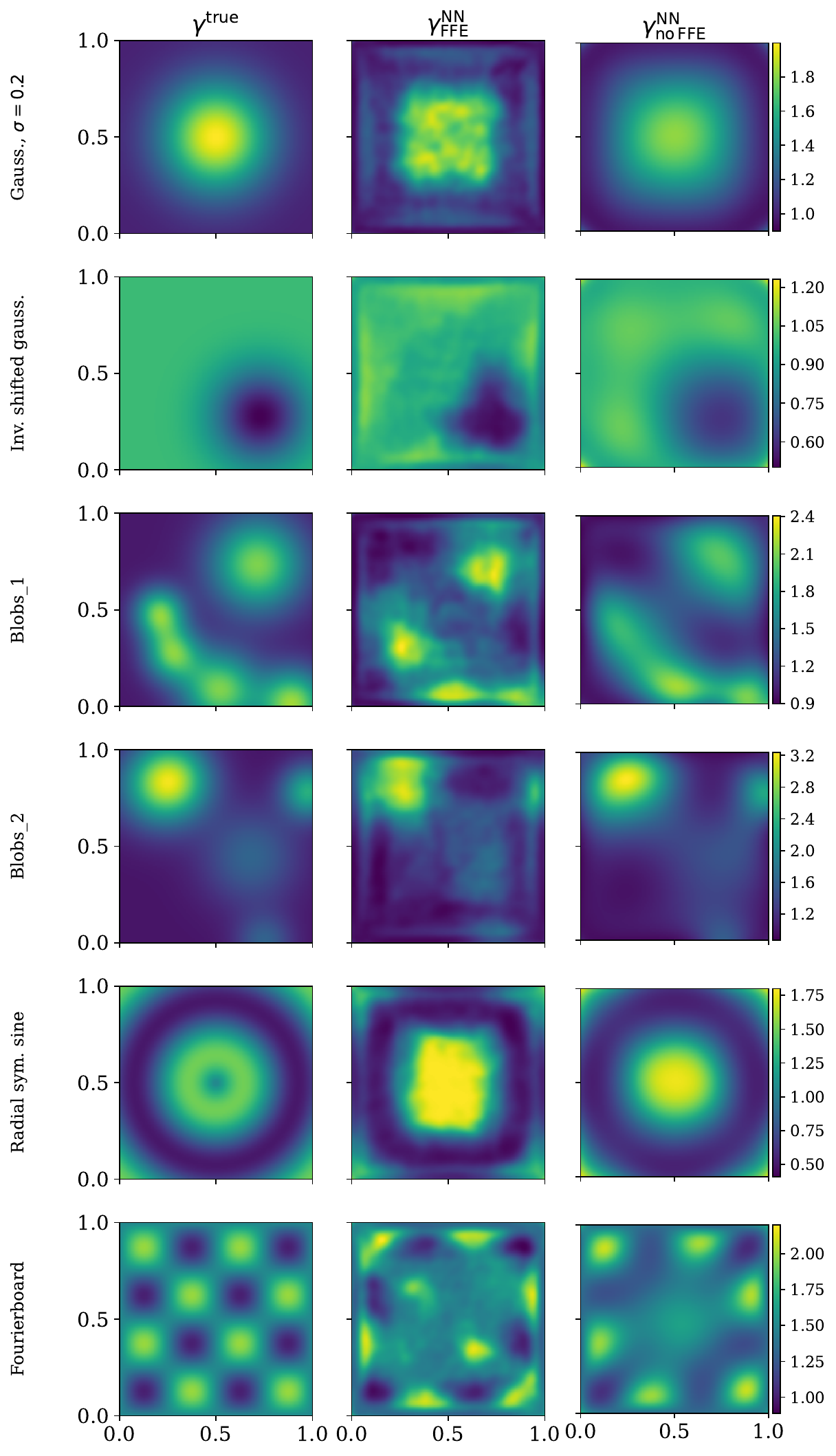}
    \caption{Reconstructed conductivities for smooth profiles. We compare the ground through (left column) with PINN reconstructions with FFE (middle column) and without FFE (right column). Each row corresponds to a different ground-truth conductivity. The conductivity/experiment label is indicated to the left side of each row.}
    \label{fig:smooth_reconstructions}
\end{figure}

We next consider smooth conductivity profiles. The corresponding reconstructions are
shown in Figure~\ref{fig:smooth_reconstructions}.


\begin{table}[t]
\centering
\resizebox{\textwidth}{!}{
\begin{tabular}{llcccccc}
\hline
\multicolumn{1}{c}{\textbf{Experiment}} &
& \multicolumn{1}{c}{\textbf{RE} $\downarrow$} 
& \multicolumn{1}{c}{\textbf{MSE} $\downarrow$} 
& \multicolumn{1}{c}{\textbf{SSIM} $\uparrow$} 
& \multicolumn{1}{c}{\textbf{IoU} $\uparrow$} 
& \multicolumn{1}{c}{\textbf{Pearson} $\uparrow$} 
& \multicolumn{1}{c}{\textbf{PSNR}}$\uparrow$ \\ 
\hline

\multirow{2}{*}{Gaussian $\sigma=0.2$}
& FFE    
& 0.061 & 0.009 & 0.815 & 0.863 & 0.940 & 20.42 \\
& no FFE 
& 0.035 & 0.003 & 0.982 & 0.831 & 0.977 & 25.13 \\
\hline

\multirow{2}{*}{Shifted Inverse Gaussian}
& FFE    
& 0.044 & 0.002 & 0.833 & 0.881 & 0.931 & 20.31 \\
& no FFE 
& 0.032 & 0.001 & 0.967 & 0.967 & 0.964 & 23.43 \\
\hline

\multirow{2}{*}{Blobs Case 1}
& FFE    
& 0.069 & 0.016 & 0.798 & 0.831 & 0.934 & 19.63 \\
& no FFE 
& 0.046 & 0.008 & 0.946 &	0.875 &	0.963 &	22.33 \\
\hline

\multirow{2}{*}{Blobs Case 2}
& FFE    
& 0.068 & 0.014 & 0.908 & 0.740 & 0.970 & 25.41 \\
& no FFE 
& 0.037 &	0.005 &	0.988 &	0.760 &	0.989 &	29.99 \\
\hline

\multirow{2}{*}{Radially Symmetric Sine}
& FFE    
& 0.115 & 0.024 & 0.827 & 0.878 & 0.953 & 16.26 \\
& no FFE 
& 0.076 & 0.010 & 0.928 & 0.893 & 0.957 & 20.01 \\
\hline

\multirow{2}{*}{FourierBoard}
& FFE    
& 0.084 & 0.025 & 0.700 & 0.616 & 0.785 & 15.96 \\
& no FFE 
& 0.082 & 0.027 & 0.810 & 0.596 & 0.751 & 15.60 \\
\hline

\end{tabular}}
\caption{Comparison of reconstruction metrics between FFE and no-FFE models for the second set of conductivity profiles. Upward arrows indicate that larger values of the metric correspond to better reconstructed profiles, while downward arrows indicate the opposite}
\label{tab:metrics_table_2}
\end{table}

For the smooth profiles considered, the raw-coordinate network performs better than the FFE model. This is consistent with the fact that these conductivities are dominated by smooth, low-frequency structure, which standard MLPs are naturally biased toward representing. In the Gaussian, shifted inverse Gaussian, and radial sine cases, the no-FFE model achieves lower relative error and MSE, and higher SSIM and PSNR. The \textit{FourierBoard} case is more mixed. Although the no-FFE model gives slightly better relative error and SSIM, the FFE model gives slightly lower MSE, higher IoU, and higher Pearson correlation. 

\subsection{Random and heterogeneous conductivities}
\begin{figure}[htbp]
    \centering
    \includegraphics[width=0.9\textwidth, keepaspectratio]{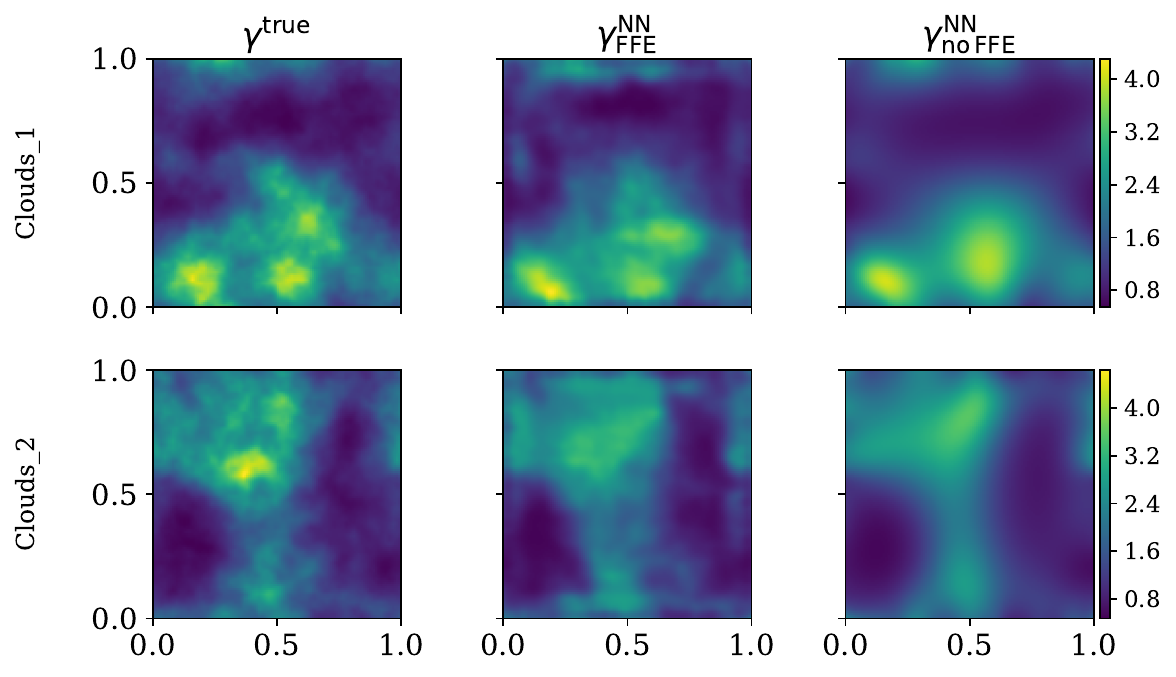}
    \caption{Reconstructed conductivities for heterogeneous profiles with random Gaussian noise in the frequency domain, denoted ``\textit{random clouds}''. We compare the ground through (left column) with PINN reconstructions with FFE (middle column) and without FFE (right column).}
    \label{fig:random_reconstructions}
\end{figure}

We finally consider random heterogeneous conductivity profiles shown in figure \ref{fig:random_reconstructions}.


\begin{table}[htbp]
\centering
\resizebox{\textwidth}{!}{
\begin{tabular}{llcccccc}
\hline
\multicolumn{1}{c}{\textbf{Experiment}} &
& \multicolumn{1}{c}{\textbf{RE} $\downarrow$} 
& \multicolumn{1}{c}{\textbf{MSE} $\downarrow$} 
& \multicolumn{1}{c}{\textbf{SSIM} $\uparrow$} 
& \multicolumn{1}{c}{\textbf{IoU} $\uparrow$} 
& \multicolumn{1}{c}{\textbf{Pearson} $\uparrow$} 
& \multicolumn{1}{c}{\textbf{PSNR}} $\uparrow$ \\ 
\hline

\multirow{2}{*}{Clouds Case 1}
& FFE    
& 0.118 & 0.059 & 0.859 & 0.862 & 0.958 & 23.60 \\
& no FFE 
& 0.103 & 0.074 & 0.893 & 0.880 & 0.948 & 22.58 \\
\hline

\multirow{2}{*}{Clouds Case 2}
& FFE    
& 0.116 & 0.078 & 0.848 & 0.849 & 0.938 & 23.63 \\
& no FFE 
& 0.100 & 0.086 & 0.890 & 0.878 & 0.932 & 23.17 \\
\hline

\end{tabular}}
\caption{Comparison of reconstruction metrics between FFE and no-FFE models for the third set of conductivity profiles. Upward arrows indicate that larger values of the metric correspond to better reconstructed profiles, while downward arrows indicate the opposite}
\label{tab:metrics_table_3}
\end{table}

The \textit{random cloud} cases in Fig. \ref{fig:random_reconstructions} show an intermediate behavior between the fully discontinuous profiles of Fig. \ref{fig:inclusions_sharp_reconstructions} (for which the power at high-frequencies is non-negligible) and the smooth, low-frequency dominated profiles in Fig. \ref{fig:smooth_reconstructions}; the spectrally-filtered Gaussian random noise used to generate these profiles populates the Fourier spectrum of the conductivity profile with both high and low frequencies (see \ref{ap:random_profiles}), making it an intermediate testing regime for the pipeline with and without FFE. 

The no-FFE model gives lower RE and higher SSIM and IoU, whereas the FFE model gives lower MSE and slightly higher Pearson correlation. Visual inspection also suggests that FFE retains more of the fine-scale structure, while the raw-coordinate reconstructions are smoother. These results indicate that FFE can improve the recovery of high-frequency, heterogeneous details without necessarily improving every global metric.

\subsection{Analysis of the power spectra with and without FFE}

\begin{figure}
    \centering
    \includegraphics[width=0.99\linewidth]{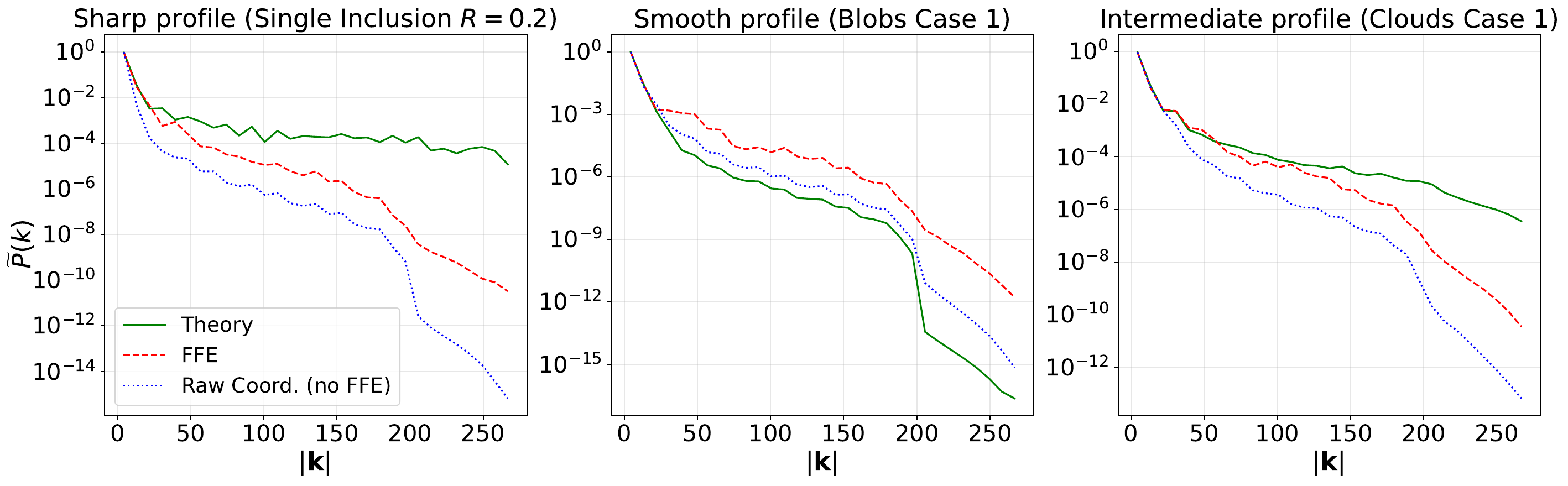}
    \caption{Comparison of scale-dependent, radially averaged, normalized power spectra $\tilde{P}(k)$ for representative examples of reconstructions of sharp, high-frequency dominated profiles (left), smooth, low-frequency dominated profiles (middle) and intermediate profiles with random gaussian frequencies (right). We show the power spectrum for the theoretical (solid green), the recovered with FFE (dashed red) and with raw coordinates (no FFE, dotted blue) profiles.}
    \label{fig: representative power spectra}
\end{figure}

We report the radially averaged power spectra of three conductivity profiles that are representative of each of the analyzed profile types i.e. sharp, smooth, and intermediate (random gaussian frequencies), see Fig. \ref{fig: representative power spectra}.

To analyze the frequency content of the reconstructed conductivity fields, we use a discrete cosine transform (DCT). Unlike the standard Fourier transform, the DCT uses a cosine basis corresponding to a symmetric extension of the field, avoiding artificial boundary discontinuities caused by non-periodic conductivity profiles. This provides a decomposition into frequency scales while preserving boundary features. For our $N$-discrete profiles:

\begin{equation}
\hat{\gamma}_{mn}
=
\alpha_m \alpha_n
\sum_{i=0}^{N-1}
\sum_{j=0}^{N-1}
\gamma_{ij}
\cos\left(
\frac{\pi}{N}\left(i+\frac{1}{2}\right)m
\right)
\cos\left(
\frac{\pi}{N}\left(j+\frac{1}{2}\right)n
\right),
\end{equation}
where $m,n=0,\ldots,N-1$ and the orthonormal normalization factors are $\alpha_0=\sqrt{1/N}, \ \alpha_{m>0}=\sqrt{2/N}$. Frequencies are $k_x(m)=\frac{\pi m}{L}$ and $k_y(n)=\frac{\pi n}{L}$.\\

The radially averaged power spectrum is computed as
\begin{equation}
P(k_q)
=
\sum_{(m,n)\in B_q}
|\hat{\gamma}_{mn}|^2 \,,
\end{equation}
where we defined radial bins as
\begin{equation}
B_q=
\left\{
(m,n):
k_q \leq k_{mn}<k_{q+1}
\right\} \qquad ; \qquad k_{mn}=\sqrt{k_x(m)^2 + k_y(n)^2}\,.
\end{equation}

Finally, the normalized power spectrum is defined as
\begin{equation}
\tilde{P}(k_q)
=
\frac{
\sum_{(m,n)\in B_q}
|\hat{\gamma}_{mn}|^2
}{
\sum_{m,n}|\hat{\gamma}_{mn}|^2}.
\end{equation}

By inspecting Figure \ref{fig: representative power spectra}, one can see that the sharp profile requires non-negligible power\footnote{In the case of piece-wise discontinuities, the power spectrum decays slowly with frequency, i.e. $P(k)\sim k^{-2}$.} from all frequencies to recover the sharp discontinuity (up to the resolution given by the $N\times N$ grid). We see that the FFE profile contains more power in the higher frequencies than the no-FFE, therefore allowing for a better reconstruction. Conversely, one can see that the smooth profile's power spectrum is better captured by the no-FFE reconstruction; the NN with FFE attempts to reduce the populated high frequencies in the high-dimensional re-parametrized space $\Phi(\mathbf{x})$, but does not reach the accuracy of the representation in raw coordinates $\mathbf{x}$. Finally, one can see that the FFE case also captures a larger portion of the higher frequencies in the intermediate profile (with frequencies sampled from a random gaussian distribution) than the no-FFE, as expected. In this sense, sharp features and high-frequencies in the conductivity are better reconstructed with the FFE method presented in this work, showing that the characteristic spectral bias of PINNs \cite{spectral_bias_Rahaman_2019} can be tackled in this problem through a combined PINN-FFE approach.

\subsection{Probing the sensitivity of the model to the center of the domain}
\label{subsec 5: center sensitivity}

We further analyze the sensitivity of our method to conductivity variations near the center of the domain. It is well known that the Calder\'on inverse problem is severely unstable when reconstructing $\gamma$ deeper inside the domain, since BCs have limited ability to probe the central region. As discussed in Sec. \ref{subsec: data generation}, we use compositions of wavelets as Dirichlet BCs $f_k$, motivated by the improved reconstruction performance observed experimentally.

\begin{figure}[t]
    \centering
    \begin{subfigure}[b]{0.49\textwidth}
        \includegraphics[width=\textwidth]{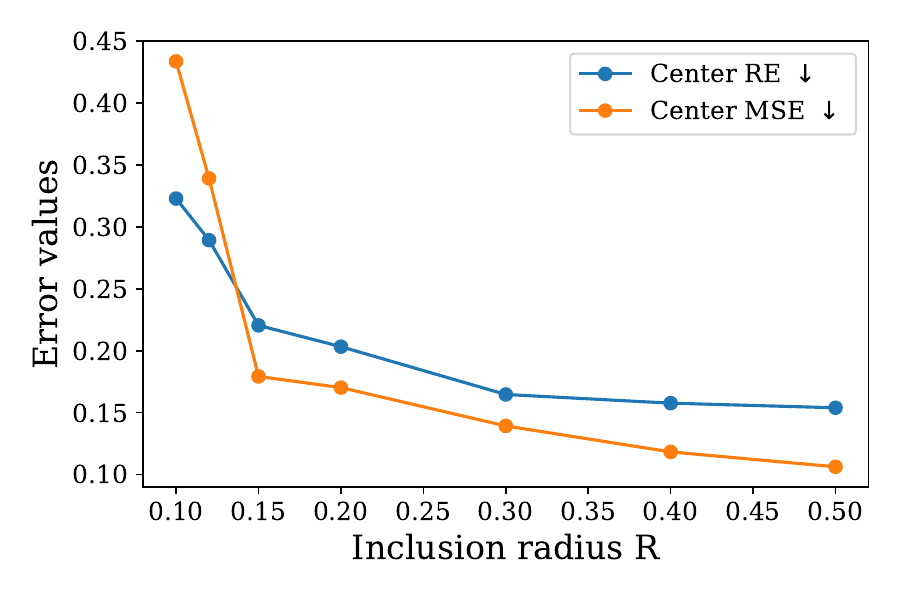}
        \caption{}
        \label{subfig: re and mse vs R}
    \end{subfigure}
    \begin{subfigure}[b]{0.49\textwidth}
        \includegraphics[width=\textwidth]{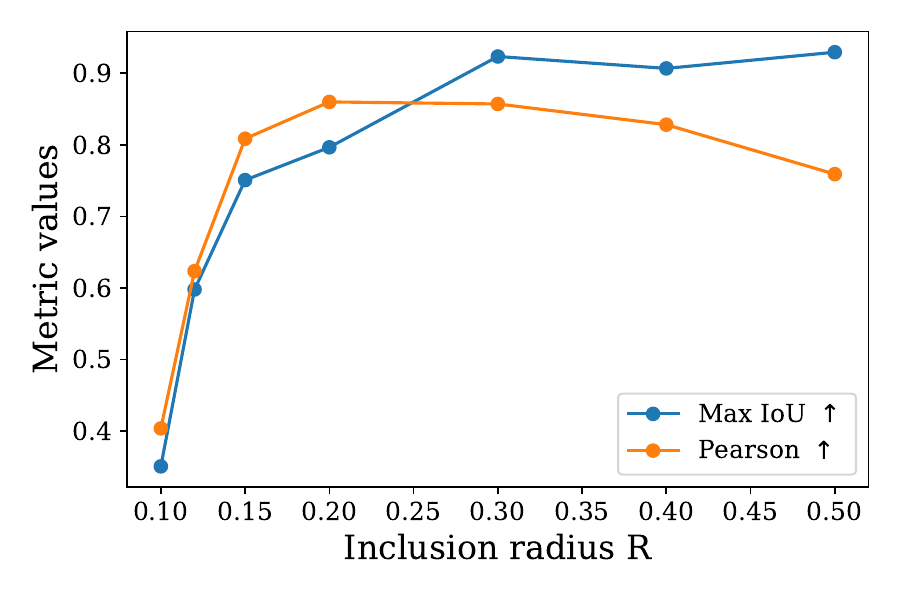}
        \caption{}
        \label{subfig: max Iou and Pearson vs R}
    \end{subfigure}
    \caption{Metrics for the recovery of the conductivity $\gamma^\text{NN}(\mathbf{x})$ for experiments with a single circular inclusion at the center of the domain with varying radius $R$, while keeping fixed BCs across all different experiments. \textbf{(a)}: RE and MSE between $\gamma^\text{NN}$ and $\gamma^\text{true}$ in the central region $\Omega_\text{center}$, with varying side-length $L_\text{center}=2R$. \textbf{(b)}: Maximum IoU value and Pearson correlation coefficient between $\gamma^\text{NN}$ and $\gamma^\text{true}$ in the full domain $\Omega$.}
    \label{fig: varying R constant K}
\end{figure}

To study this effect, we perform the following sensitivity analysis. We fix a single realization of the randomized BCs $f_k$, with $k=~1,\dots,140$, as well as the random initialization of the NN pipeline and the number of training epochs. We then consider seven experiments with a single circular inclusion located at the center of the domain, varying only its radius $R$. We run the inversion PINN pipeline with the FFE method. The corresponding results are shown in Fig. \ref{fig: varying R constant K}, where we report the RE, MSE, maximum IoU, and Pearson correlation coefficient as functions of $R$. The point-wise metrics RE and MSE shown in Fig. \ref{subfig: re and mse vs R} are computed only within a near-center subdomain $$\Omega_\text{center}=\left(\frac{L}{2}-R, \, \frac{L}{2} + R\right) \times \left(\frac{L}{2}-R, \, \frac{L}{2} + R\right)\ .$$
This is a square located at the center of $\Omega$, with side length equal to two times the inclusion radius, $L_\text{center}=2R$. This normalization is necessary because computing RE and MSE over the full domain would artificially favor smaller inclusions: as the inclusion size decreases, a larger portion of the domain corresponds to the constant background conductivity $\gamma_0=1$, leading to smaller average errors even for poor reconstructions, such as a constant conductivity with value $\gamma_0$ over the whole domain. Restricting the evaluation to $\Omega_\text{center}$ ensures that the relative proportion between inclusion and background remains constant across experiments, allowing for a more meaningful assessment of reconstruction quality near the center of the domain.

\begin{figure}[t]
    \centering
    \begin{subfigure}[b]{0.49\textwidth}
        \includegraphics[width=\textwidth]{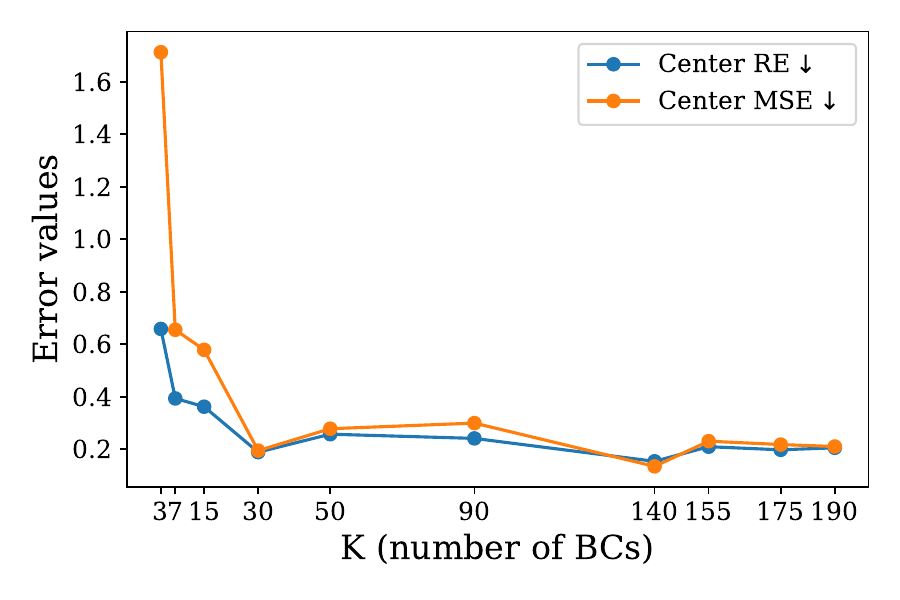}
        \caption{}
        \label{subfig: re and mse vs K}
    \end{subfigure}
    \begin{subfigure}[b]{0.49\textwidth}
        \includegraphics[width=\textwidth]{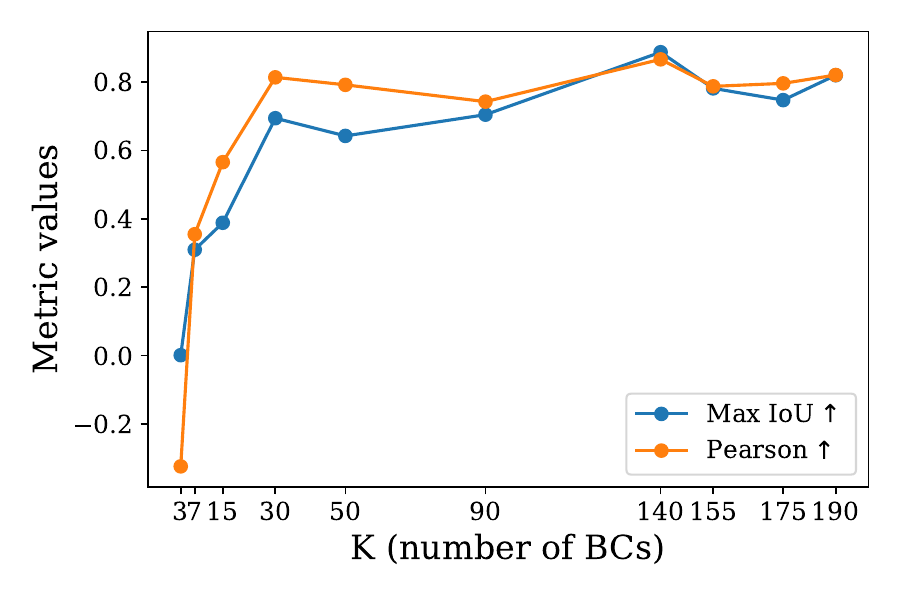}
        \caption{}
        \label{subfig: max Iou and Pearson vs K}
    \end{subfigure}
    \caption{Metrics for the recovery of the conductivity $\gamma^\text{NN}(\mathbf{x})$ for experiments with a single circular inclusion at the center of the domain with fixed radius $R=0.2L$, for varying number of BCs across all different experiments. \textbf{(a)}: RE and MSE between $\gamma^\text{NN}$ and $\gamma^\text{true}$ in the central region $\Omega_\text{center}$, with side-length proportional to $R$, $L_\text{center}=2R$. \textbf{(b)}: Maximum IoU value and Pearson correlation coefficient between $\gamma^\text{NN}$ and $\gamma^\text{true}$ in the full domain $\Omega$.}
    \label{fig: varying K constant R}
\end{figure}

From Fig. \ref{fig: varying R constant K}, we observe that both RE and MSE increase as the inclusion radius decreases, indicating that recovering the conductivity near the center becomes progressively more difficult for smaller inclusions. Consistently, the Pearson correlation coefficient and the maximum IoU decrease as $R$ decreases, further supporting this experimentally observed loss of reconstruction quality in the central region.

Additionally, we performed the complementary analysis: we fix the radius of the single inclusion at the center of the domain to $R=0.2L$, and vary the amount of DtN data pairs $\{f_k, J_k\}_{k=1}^K$ used in the reconstruction, i.e. we vary $K$ while keeping $R$ constant. Note that the BCs are randomly sampled and the DtN data is generated only once, according to procedure described in Sec. \ref{subsec: data generation}. The number of data pairs generated is the highest value of K in the ten experiments, in this case $K=190$. Then, the PINN reconstruction is performed for this value of K. Later, the initial number of pairs is reduced to a subset of them, and the pipeline is re-run. This is done recursively up to the minimum amount of data pairs displayed in Fig. \ref{fig: varying K constant R} (i.e. $K=3$). One can see that decreasing the number of DtN data pairs used in the reconstruction worsens the recovery metrics, although the pattern is less significant up until a number of data pairs around $K=30$. For lower $K$, the reconstruction quality worsens significantly, providing a reference threshold for the minimum amount of pairs needed to reach a certain reconstruction quality.

\section{Conclusions and Future Directions}
\label{sec 6: conclusions}

We developed a PINN framework for reconstructing a spatially varying
conductivity from a finite collection of boundary voltage--current
measurements, or DtN pairs. The method uses one NN to represent the
conductivity and a second conditional NN to represent the family of
electric potentials associated with the imposed boundary excitations. The
two networks are trained jointly using the governing elliptic PDE and the
available Dirichlet and Neumann data.

The main objective was to study how boundary-probe design and coordinate
representation affect finite-data reconstruction. Randomized wavelet
superpositions provide localized multiscale boundary excitations, while
FFE modifies the spectral bias of the neural
representation. For localized inclusion-type profiles, FFE generally
improves point-wise accuracy and geometric overlap. Its advantage is not
universal, however: raw-coordinate networks perform comparably or better
for several smooth and low-frequency dominated conductivity profiles.

The experiments also demonstrate the limitations of the reconstruction.
Smaller inclusions near the center of the domain are more difficult to
recover, and different evaluation metrics can favor different aspects of
the reconstructed field. These observations are consistent with the
intrinsic instability of the finite-data Calder\'on inverse problem and show that
the recovered conductivity should be interpreted as a reconstruction
consistent with the available measurements and the implicit regularization
of the neural parametrization.

The present study is restricted to synthetic data generated using the same mathematical model imposed during training. Future work should examine measurement noise, modeling errors, electrode effects, partial boundary data, irregular geometries, and uncertainty quantification. It would also be useful to compare the randomized wavelet probes systematically with alternative or adaptively optimized boundary excitations.

Overall, the results show that PINN-based inversion provides a flexible framework for finite-data conductivity reconstruction. FFEs
are particularly useful for localized sharp structures, but the preferred coordinate representation depends on the spatial and frequency character of the conductivity being reconstructed. In particular, we have found that a non-negligible amount of power contained in the high-frequency modes is one of the relevant features for which the FFE method improves the reconstruction.

\section*{CRediT authorship contribution statement}

AAK and PTe contributed to code development, numerical experiment design, data generation, formal analysis, and visualization. ME and GD contributed with mathematical expertise and guidance on the Calderón inverse conductivity problem. All authors contributed to the conceptualization and methodology of the work, the analysis and interpretation of the results, and the writing, editing, and review of the manuscript.

\section*{Declaration of competing interest}
The authors declare that they have no known competing financial interests or personal relationships that could have appeared to influence the work reported in this paper.

\section*{Data availability}
Codebase will be released in Github after acceptance.
\section*{Acknowledgements}

This work was supported by a grant from the Simons Foundation (00017375, RJ, GD, ME).
Funding for the work of RJ, AAK, KK, LS, PTa and PTe was partially provided by project PID2022-141125NB-I00,
and the “Center of Excellence Maria de Maeztu 2025-2029” award to the ICCUB funded by
grant CEX2024-001451-M from AEI/10.13039/501100011033.
PTa is supported by the project “Dark Energy and the Origin of the Universe” (PRE2022-102220), funded by MCIN/AEI/10.13039/501100011033. Funding for the work of ME was also partially supported by  NSF DMS CAREER 2143719.

\bibliographystyle{elsarticle-num}

\newpage

\appendix

\section{Finite Differences Method (FDM) forward solver}
\label{appendix:fdm solver}
To train and evaluate the aforementioned method, we generated synthetic boundary data by solving the forward problem numerically. The governing equation \eqref{eq: calderon ODE}:
\begin{equation}
\nabla \cdot (\gamma(x)\nabla u(x)) = 0
\end{equation}
on the square domain $\Omega=[0,1]\times[0,1]$, subject to prescribed Dirichlet BCs
\begin{equation}
u|_{\partial\Omega}=f.
\end{equation}
For each chosen conductivity profile $\gamma(x)$ and boundary excitation $f$, the corresponding potential field $u$ was computed using a finite-difference discretization on a uniform $N\times N$ grid. Let
\begin{equation}
x_j=jh,\qquad y_i=ih,\qquad h=\frac{1}{N-1},
\end{equation}
with $i,j=0,\dots,N-1$. The conductivity and potential are represented at the grid nodes as
\begin{equation}
\gamma_{i,j}\approx \gamma(x_j,y_i),\qquad u_{i,j}\approx u(x_j,y_i).
\end{equation}
We directly discretize the divergence-form operator. At each interior node $(i,j)$, the equation is approximated by the finite-difference scheme
\begin{equation}
\begin{aligned}
0=\frac{1}{h^2}\Big[
&\gamma_{i,j+\frac12}(u_{i,j+1}-u_{i,j})
-\gamma_{i,j-\frac12}(u_{i,j}-u_{i,j-1})\\
&+\gamma_{i+\frac12,j}(u_{i+1,j}-u_{i,j})
-\gamma_{i-\frac12,j}(u_{i,j}-u_{i-1,j})
\Big].
\end{aligned}
\end{equation}
This corresponds to a second-order discretization of the fluxes across the interfaces between neighboring grid points.

\noindent Since the conductivity is only known at the nodes, interface values such as $\gamma_{i,j+\frac12}$ must be approximated from neighboring nodal values. In this work, these interface conductivities are computed using harmonic averaging, for example
\begin{equation}
\gamma_{i,j+\frac12}
=
\frac{2\gamma_{i,j}\gamma_{i,j+1}}{\gamma_{i,j}+\gamma_{i,j+1}},
\end{equation}
and analogously for the west, north, and south interfaces.

After discretization, the forward problem becomes a sparse linear system $A\mathbf{u}=\mathbf{b}$. Interior rows correspond to the finite-difference stencil of the PDE, while dirichlet BCs were imposed directly on the boundary nodes of the grid. The sparse linear system was solved using SciPy's routine \texttt{spsolve}. Once the potential was obtained, the boundary current density was computed from
\begin{equation}
    J_\gamma(f)=\left[\gamma(x)\frac{\partial u}{\partial\hat{n}}\right]_{\partial\Omega}
\end{equation}
The normal derivative at the boundary was approximated using second-order one-sided finite differences. For example, at the left boundary one has
\begin{equation}
\frac{\partial u}{\partial x}(0,y_i)
\approx
\frac{-3u_{i,0}+4u_{i,1}-u_{i,2}}{2h},
\end{equation}
with analogous expressions at the other sides. The normal derivative was then obtained by projecting the gradient onto the outward unit normal vector,
\begin{equation}
\frac{\partial u}{\partial n} = \nabla u \cdot n.
\end{equation}
In this way, for each chosen boundary excitation $f_k$, the forward problem is solved to obtain the corresponding boundary current $J_k=\Lambda_\gamma(f_k)$. Each simulation therefore provides one DtN pair $(f_k,J_k)$.

\section{Conductivity profiles used in the numerical experiments}
\label{ap:conductivity_profiles}

In this appendix we specify the ground-truth conductivity profiles used in the numerical experiments of Sec.~\ref{sec: results}. All profiles are defined on the unit square
\[
\Omega=[0,1]\times[0,1].
\]

\subsection{Sharp and inclusion-type conductivities}
\label{ap:sharp_profiles}

\begin{itemize}

    \item \textbf{Single circular inclusion.}
    The conductivity is defined by
    \begin{equation}
        \gamma^{\mathrm{true}}(\mathbf{x};R,\mathbf{c},\gamma_0,\gamma_1)
        =
        \begin{cases}
            \gamma_1, & (x-c_x)^2+(y-c_y)^2 \leq R^2, \\[4pt]
            \gamma_0, & (x-c_x)^2+(y-c_y)^2 > R^2,
        \end{cases}
        \label{eq:single_inclusion_profile}
    \end{equation}
    where \(\mathbf{x}=(x,y)\), \(\mathbf{c}=(c_x,c_y)\), \(R\) is the inclusion radius, \(\gamma_0\) is the background conductivity, and \(\gamma_1\) is the inclusion conductivity. We use
    \[
    \mathbf{c}=\left(\frac{1}{2},\frac{1}{2}\right),
    \qquad
    \gamma_0=1,
    \qquad
    \gamma_1=2,
    \]
    with two radii, \(R=0.2\) and \(R=0.15\).

    \item \textbf{Two circular inclusions.}
    The conductivity is defined by
    \begin{equation}
        \gamma^{\mathrm{true}}(\mathbf{x})
        =
        \begin{cases}
            \gamma_1, & (x-c_{x,1})^2+(y-c_{y,1})^2 \leq R_1^2, \\[4pt]
            \gamma_2, & (x-c_{x,2})^2+(y-c_{y,2})^2 \leq R_2^2, \\[4pt]
            \gamma_0, & \mathrm{otherwise}.
        \end{cases}
        \label{eq:two_inclusions_profile}
    \end{equation}
    The parameters are
    \[
    \gamma_0=1,\qquad
    R_1=0.2,\qquad
    \mathbf{c}_1=\left(\frac{1}{2},\frac{1}{2}\right),
    \qquad
    \gamma_1=\frac{3}{2},
    \]
    and
    \[
    R_2=0.15,\qquad
    \mathbf{c}_2=\left(\frac{1}{5},\frac{4}{5}\right),
    \qquad
    \gamma_2=2.25.
    \]

    \item \textbf{Piecewise-constant horizontal split.}
    The conductivity is defined by
    \begin{equation}
        \gamma^{\mathrm{true}}(x,y)
        =
        \begin{cases}
            \gamma_{\mathrm{bottom}}, & y < \frac{1}{2}, \\[4pt]
            \gamma_{\mathrm{top}}, & y \geq \frac{1}{2}.
        \end{cases}
        \label{eq:piecewise_constant_profile}
    \end{equation}
    with
    \[
    \gamma_{\mathrm{bottom}}=0.5,
    \qquad
    \gamma_{\mathrm{top}}=2.
    \]

\end{itemize}

\subsection{Smooth conductivities}
\label{ap:smooth_profiles}

\begin{itemize}

    \item \textbf{Gaussian profile.}
    The conductivity is
    \begin{equation}
        \gamma^{\mathrm{true}}(\mathbf{x})
        =
        \gamma_0
        +
        \exp\left(
        -\frac{\|\mathbf{x}-\mathbf{c}\|^2}{2\sigma^2}
        \right),
        \label{eq:gaussian_profile}
    \end{equation}
    with
    \[
    \gamma_0=1,
    \qquad
    \mathbf{c}=\left(\frac{1}{2},\frac{1}{2}\right),
    \qquad
    \sigma=0.2.
    \]

    \item \textbf{Shifted inverse Gaussian profile.}
    The conductivity is
    \begin{equation}
        \gamma^{\mathrm{true}}(\mathbf{x})
        =
        \gamma_0
        -
        d\exp\left(
        -\frac{\|\mathbf{x}-\mathbf{c}\|^2}{2\sigma^2}
        \right),
        \label{eq:shifted_inverse_gaussian_profile}
    \end{equation}
    with
    \[
    \gamma_0=1,
    \qquad
    d=0.5,
    \qquad
    \mathbf{c}=(0.73,0.28),
    \qquad
    \sigma=0.17.
    \]

    \item \textbf{Random Gaussian blobs.}
    The conductivity is constructed as
    \begin{equation}
        \gamma^{\mathrm{true}}(\mathbf{x})
        =
        \gamma_0
        +
        \sum_{i=1}^{N_{\mathrm{blobs}}}
        A_i
        \exp\left(
        -\frac{\|\mathbf{x}-\mathbf{c}_i\|^2}{2\sigma_i^2}
        \right).
        \label{eq:random_blobs_profile}
    \end{equation}
    We use
    \[
    \gamma_0=1,
    \qquad
    N_{\mathrm{blobs}}=5,
    \]
    with parameters sampled independently as
    \begin{equation}
        A_i \sim \mathcal{U}([0.5,1.5]),
        \qquad
        \mathbf{c}_i \sim \mathcal{U}(\mathrm{int}(\Omega)),
        \qquad
        \sigma_i \sim \mathcal{U}([0.015,0.2]).
        \label{eq:random_blobs_parameters}
    \end{equation}
    The two blob cases correspond to two different random draws.
    
    \item \textbf{Radially symmetric sinusoidal profile.}
    The conductivity is
    \begin{equation}
        \gamma^{\mathrm{true}}(\mathbf{x})
        =
        1 + A\sin\left(2\pi f r(\mathbf{x})\right),
        \qquad
        r(\mathbf{x})=\|\mathbf{x}-\mathbf{c}\|,
        \label{eq:radial_sine_profile}
    \end{equation}
    with
    \[
    A=0.5,
    \qquad
    f=1.75,
    \qquad
    \mathbf{c}=(0.5,0.5).
    \]

    \item \textbf{\textit{FourierBoard} profile.}
    The conductivity is
    \begin{equation}
        \gamma^{\mathrm{true}}(x,y)
        =
        \frac{3}{2}
        +
        A\sin\left(\frac{m\pi x}{L}\right)
        \cos\left(\frac{n\pi y}{L}+\frac{\pi}{8}\frac{n}{L}\right),
        \label{eq:fourierboard_profile}
    \end{equation}
    with
    \[
    m=4,
    \qquad
    n=4,
    \qquad
    L=1,
    \qquad
    A=\frac{1}{2}.
    \]

\end{itemize}

\subsection{Random and heterogeneous conductivities}
\label{ap:random_profiles}

\begin{itemize}

    \item \textbf{Spectrally filtered random clouds.}
    The conductivity is generated as a spectrally filtered Gaussian random field:
    \begin{equation}
        \gamma^{\mathrm{true}}(\mathbf{x})
        =
        \gamma_0
        +
        A\,\mathcal{F}^{-1}
        \left[
        \hat{\gamma}(\mathbf{k})
        \right](\mathbf{x}),
        \label{eq:random_clouds_profile}
    \end{equation}
    where \(\mathbf{k}=(k_1,k_2)\in\mathbb{Z}^2\), and
    \begin{equation}
        \hat{\gamma}(\mathbf{k})
        =
        \frac{\hat{\xi}(\mathbf{k})}
        {\left(1+\|\mathbf{k}\|^2\right)^{\alpha/2}},
        \qquad
        \|\mathbf{k}\|^2=k_1^2+k_2^2.
        \label{eq:random_clouds_spectrum}
    \end{equation}
    The random Fourier amplitudes are sampled as
    \begin{equation}
        \hat{\xi}(\mathbf{k})
        \sim
        \mathcal{N}(0,1)
        +
        i\,\mathcal{N}(0,1),
        \label{eq:random_clouds_noise}
    \end{equation}
    with Hermitian symmetry imposed to ensure that \(\gamma^{\mathrm{true}}\) is real-valued. We use
    \[
    \alpha=2,
    \qquad
    \gamma_0=\frac{3}{2},
    \qquad
    A=\frac{1}{2}.
    \]
    The two cloud cases correspond to two different random draws of the Fourier-space noise field.

\end{itemize}

\section{Training specifications and loss values for all experiments} \label{appendix: training and loss specifications}

In this appendix, we give specifications on the training and architecture details of the PINNs, and report on the loss function values obtained for all experiments (see Table \ref{tab:training_losses}).

Unless stated otherwise, all experiments were performed on the square domain $\Omega=[0,1]^2$ using uniformly sampled interior collocation points and synthetic boundary datasets generated with the finite-difference forward solver described previously. For a grid resolution $N\times N$, the boundary dataset contains $4N-4$ boundary nodes per imposed BC. In the typical setup used throughout this work, $N=64$, yielding $252$ boundary nodes per excitation.

At each optimization step, the PDE residual was evaluated on randomly sampled interior collocation points ($N_{\mathrm{PDE}}=4096$), while boundary samples from the dataset were processed in shuffled mini-batches of size $N_{\mathrm{bnd}}=1024$. The potential network $u^{\mathrm{NN}}$ was conditioned on the boundary-condition index through a one-hot  encoding concatenated to the spatial coordinates (or their Fourier-feature representation), allowing a single shared network to represent the full family of potentials $\{u_k\}_{k=1}^K$.

The conductivity network consisted of $4$ hidden layers with $128$ neurons each, while the potential network used $6$ hidden layers of $256$ neurons. Both employed SiLU or ReLu activations. The conductivity output was constrained to a physically admissible interval through a sigmoid reparametrization where $\gamma_{min} = 0.5$ and $\gamma_{max} = 2.5, 10$, depending on the conductivity profile. Training was performed using the \textit{Adam} optimizer with learning rate $\eta=10^{-4}$ together with an exponential learning-rate scheduler.

\begin{figure}[htbp]
    \centering

    \begin{subfigure}[t]{0.49\linewidth}
        \centering
        \includegraphics[width=\linewidth]
        {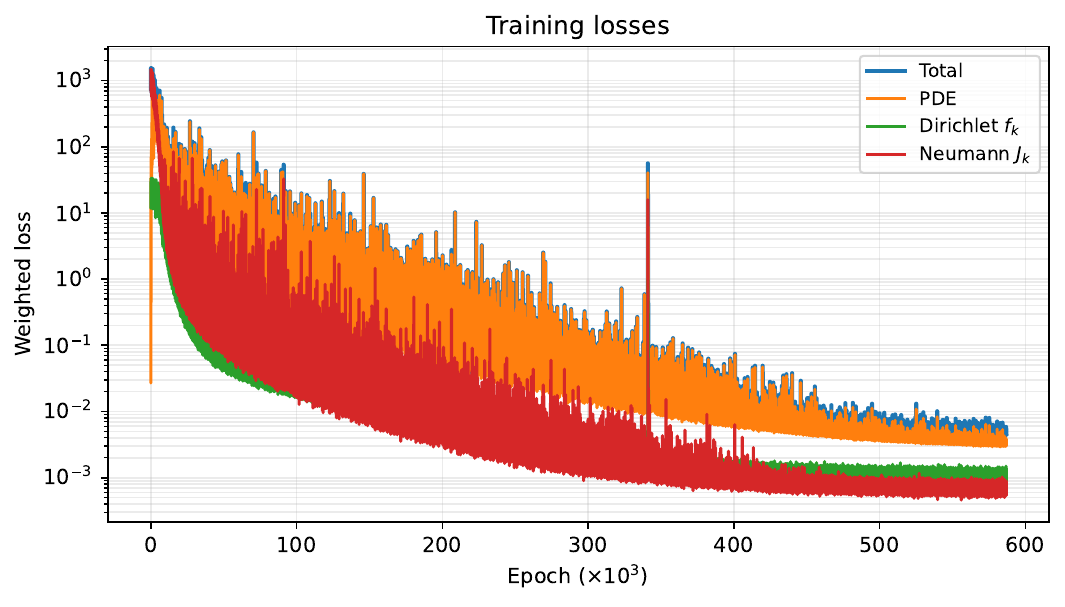}
        \caption{Training losses with FFE.}
        \label{fig:loss-radial-sine-full}
    \end{subfigure}
    \hfill
    \begin{subfigure}[t]{0.49\linewidth}
        \centering
        \includegraphics[width=\linewidth]
        {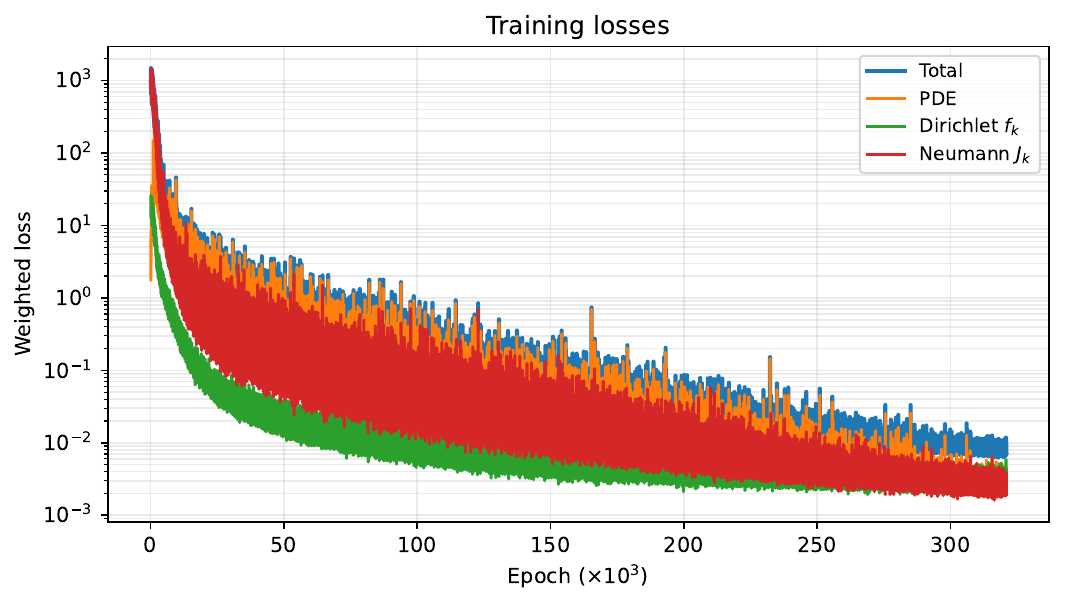}
        \caption{Training losses without FFE.}
        \label{fig:loss-radial-sine-truncated}
    \end{subfigure}

    \caption{
        Representative training-loss evolution for the two-inclusion profile with and without FFE.
    }
    \label{fig:loss-radial-sine}
\end{figure}

We present here the graphs of representative loss functions (for the two-inclusions profile of Fig. \ref{fig:loss-radial-sine}), and the table containing loss values alongside epochs trained for all experiments, with and without FFE. Training was continued until the total loss and its main components reached an approximately stable regime. We do not display all loss-plots, since they all share a general decreasing trend like the example of Fig. \ref{fig:loss-radial-sine}.

\begin{table}[htbp]
\centering
\small
\renewcommand{\arraystretch}{1.25}
\setlength{\tabcolsep}{2pt}
\caption{Final training losses for FFE and no-FFE reconstructions. Column headers indicate the power of ten by which each entry should be multiplied. All experiments are done with 140 BCs. Training was continued until the total loss and its main components reached an approximately stable regime.}
\label{tab:training_losses}

\resizebox{\textwidth}{!}{
\begin{tabular}{llccccccc}
\hline
Profile & Model & Epochs 
& Total $(10^{-2})$ 
& PDE $(10^{-2})$ 
& Dir. $(10^{-4})$ 
& Neu. $(10^{-4})$ 
& $\lambda_fL_f$ $(10^{-4})$ 
& $\lambda_JL_J$ $(10^{-3})$ \\
\hline

\multirow{2}{*}{Incl. $R=0.2$}
& FFE    & 291000 & 5.25 & 4.68 & 2.80 & 2.87 & 28.00 & 2.87 \\
& no FFE & 291000       &1.26      &  0.65    & 2.42     &  3.63   &   24.25    &  3.63    \\
\hline

\multirow{2}{*}{Incl. $R=0.15$}
& FFE    & 537000 & 0.58 & 0.40 & 0.90 & 0.87 & 8.98 & 0.87 \\
& no FFE &   241000     & 1.51     &  0.68    & 2.09     & 6.20     &   20.93   &   6.20   \\
\hline

\multirow{2}{*}{Two incl.}
& FFE    & 587000 & 0.46 & 0.32 & 0.85 & 0.63 & 8.51 & 0.63 \\
& no FFE & 321000       &0.773      &0.22      & 2.74     &  2.41    & 27.45     & 2.41     \\
\hline

\multirow{2}{*}{Gaussian}
& FFE    & 260200 & 11.87 & 9.93 & 0.09 & 10.20 & 92 & 10.20 \\
& no FFE &  311000      & 0.87      &  0.30    &  1.37    & 4.31      & 13.73   &  4.31     \\
\hline

\multirow{2}{*}{Inv. shifted gau.}
& FFE    & 161000 & 0.35 & 0.24 & 0.42 & 0.66 & 4.15 & 0.66 \\
& no FFE &  280010      & 0.27     &0.07      & 0.40     &1.57      &  4.04    & 1.57     \\
\hline

\multirow{2}{*}{Radial sym. sine}
& FFE    & 617000 & 0.81 & 0.46 & 1.52 & 1.96 & 15.20 & 1.96 \\
& no FFE & 231000       &  1.74    & 0.71     & 3.04    &  7.33     & 30.48      & 7.33      \\
\hline

\multirow{2}{*}{Piecewise cst}
& FFE    & 161000 & 1.63 & 1.10 & 3.52 & 1.75 & 35.20 & 1.75 \\
& no FFE & 290010       &  4.79    & 3.05     &  2.14    & 15.27     &  21.47     & 15.27     \\
\hline

\multirow{2}{*}{Fourierboard}
& FFE    & 400300 & 7.15 & 3.53 & 0.17 & 18.80 & 174 & 18.80 \\
& no FFE & 392000       &  8.53    & 2.02     & 0.32     &  32.58     &  325   &  32.58     \\
\hline

\multirow{2}{*}{Blob 825}
& FFE    & 400110 & 4.65 & 2.93 & 0.08 & 8.90 & 83 & 8.90 \\
& no FFE &210100        &10.14      & 4.88     & 0.16     & 35.63     & 168   &  35.63    \\
\hline

\multirow{2}{*}{Blob 886}
& FFE    & 400200 & 2.97 & 1.66 & 0.06 & 6.70 & 65 & 6.70 \\
& no FFE &  210100      & 9.71     & 2.58     & 0.29    & 41.58      & 296   & 41.58     \\
\hline

\multirow{2}{*}{Cloud 119}
& FFE    & 390030 & 20.78 & 12.04 & 0.43 & 44.0 & 431 & 44.0 \\
& no FFE & 240100       & 155      & 42.4      & 1.25     & 1007     & 1257    &  1007    \\
\hline

\multirow{2}{*}{Cloud 541}
& FFE    & 400210 & 17.90 & 8.80 & 0.48 & 42.60 & 482 & 42.60 \\
& no FFE & 210100       &  175     & 35.4     & 1.25     & 1278      & 1253    &  1278     \\
\hline

\end{tabular}}
\end{table}

\noindent Table~\ref{tab:training_losses} reports the final loss values for the FFE and no-FFE models. The number of epochs differs between runs because training was continued until the total loss and its main components reached an approximately stable regime, rather than being stopped at a fixed epoch for all experiments. The Dirichlet and Neumann losses remain small across the different profiles, indicating that the learned potentials reproduce the imposed boundary data and measured currents with good accuracy. Differences in the final PDE loss reflect the varying difficulty of representing each conductivity profile and the corresponding family of potential fields.

\end{document}